\documentclass[journal,twoside,web]{ieeecolor}
\usepackage{generic}
\usepackage{cite}
\usepackage{amsmath,amssymb,amsfonts}
\usepackage{graphicx}
\usepackage{hyperref}
\hypersetup{hidelinks=true}
\usepackage{textcomp}
\usepackage{booktabs}
\usepackage{longtable}
\usepackage{multirow}

\usepackage[lined,algonl,boxed]{algorithm2e}
\usepackage{subcaption}
\usepackage{booktabs} 
\usepackage{tabularx} 
\usepackage{bm}

\newcommand{\hide}[1]{} 

\newcommand{\RC}{ZhiFangDanTai}
\newcommand{\sRC}{ZhiFangDanTai\space}

\def\BibTeX{{\rm B\kern-.05em{\sc i\kern-.025em b}\kern-.08em
    T\kern-.1667em\lower.7ex\hbox{E}\kern-.125emX}}
\begin{document}
\title{ZhiFangDanTai: Fine-tuning Graph-based Retrieval-Augmented Generation Model for Traditional Chinese Medicine Formula}
\author{Zixuan Zhang, Bowen Hao, Yingjie Li, and Hongzhi Yin, \IEEEmembership{Senior Member, IEEE}
\thanks{Received 18 March 2025, Revised 5 June 2025, Accepted 5 Sep 2025. This work was supported by National Natural Science Foundation of China (No.62402328), R\&D Program of Beijing Municipal Education Commission (No.KM202410028007),
the Undergraduate ``Qiyan" Program of Beijing Natural Science Foundation (grant No.QY25392), the Teaching Reform Project of Capital Normal University (grant No. 2025CNUJWC030), the Digital Course Construction Project Based on Knowledge Graph of Capital Normal University in 2025, the Australian Research Council partially supports this work under the streams of Future Fellowship (Grant No. FT210100624),  the Discovery Project (Grant No. DP240101108), and the Linkage Projects (Grant No. LP230200892 and LP240200546).}
\thanks{Zixuan Zhang, Bowen Hao and Yingjie Li are with the School of Management, Capital Normal University, Beijing, China. (e-mail: 1222905049@cnu.edu.cn; 6974@cnu.edu.cn; 1222905034@cnu.edu.cn ). }
\thanks{Hongzhi Yin is with the School of Electrical Engineering and Computer Science, the University of Queensland, Brisbane,
Australia. (e-mail: h.yin1@uq.edu.au).}
\thanks{Corresponding Authors: Bowen Hao; Hongzhi Yin.}
}

\maketitle

\begin{abstract}
 Traditional Chinese Medicine (TCM) formulas play a significant role in treating epidemics and complex diseases. Existing models for TCM utilize traditional algorithms or deep learning techniques to analyze formula relationships, yet lack comprehensive results, such as complete formula compositions and detailed explanations.  Although recent efforts have used TCM  instruction datasets to fine-tune Large Language Models (LLMs) for explainable formula generation, existing datasets lack sufficient details, such as the roles of the formula's sovereign, minister, assistant, courier; efficacy; contraindications; tongue and pulse diagnosis——limiting the depth of model outputs.

To address these challenges, we propose~\RC, a  framework combining Graph-based Retrieval-Augmented Generation (GraphRAG) with LLM fine-tuning. ~\sRC uses GraphRAG to retrieve and synthesize structured TCM knowledge into concise summaries, while also constructing an enhanced instruction dataset to improve LLMs' ability to integrate retrieved information. 
Furthermore, we provide novel theoretical proofs demonstrating that integrating GraphRAG with fine-tuning techniques  can reduce generalization error and hallucination rates in the TCM formula task.
Experimental results on both collected and clinical datasets demonstrate that~\sRC achieves significant improvements over state-of-the-art models.
 Our model is open-sourced at https://huggingface.co/tczzx6/ZhiFangDanTai1.0.
\end{abstract}


\section{Introduction}
\label{sec:introduction}
\IEEEPARstart{T}raditional Chinese Medicine (TCM) has evolved over thousands of years. Its theories cover all aspects of human life and are deeply rooted in concepts like Yin-yang, the five elements, Zang-fu, meridians, Qi, blood, and body fluids. As a core component of this system, TCM formulas emphasize holistic care and personalized treatment. Through herbal medicine, acupuncture, and tuina (massage), TCM effectively addresses diverse illnesses, including epidemics and complex chronic conditions.

With the advent of the Artificial Intelligence (AI) era, AI-assisted systems for TCM formula generation have emerged to address the shortage of high-quality medical resources. As an AI-assisted tool, TCM formula systems can offer convenient medical consultation services for patients with non-complex conditions, while also supporting clinical decision-making for practitioners.
Existing AI-assisted TCM formula  models can be categorized into traditional TCM formula  models, deep learning-based TCM formula  models, and Large Language Model (LLM)-based TCM formula models. Traditional TCM formula  models employ association analysis~\cite{JingCNKI2024-1}, clustering~\cite{LiuTCM2017-2}, complex networks~\cite{NimaTCM2021-3}, and graph neural networks~\cite{ZhouPharmacol-10} to explore the associations between any two herbs in TCM formulas. Due to the limited fitting capabilities of these algorithms, such methods cannot provide complete TCM formulas. 
Deep learning-based TCM formula models aim to model the TCM diagnostic process of "wang (inspection), wen (listening and smelling), wen (inquiry), and qie (palpation)" through representation learning. 
Specifically, they first use convolutional neural networks~\cite{Zhang2024bioinformatics}, GCN~\cite{Wang2023,Lu2025Nature,Hao2021} or discriminative models~\cite{Wang2023bioinformation,Hao2019} to encode clinical indicators such as tongue coating characteristics, pulse patterns, and patient-reported symptoms; and then generate TCM formulas using established architectures including BERT~\cite{Chen2023,Hao2020ECML}, Seq2seq models~\cite{LiuAI2022-21}, or GAN~\cite{Wang2023bioinformation}, with some approaches incorporating reinforcement learning~\cite{Wang2023} to optimize the recommendation quality.
Although these methods can provide complete TCM formulas, they fail to provide explainable reasons. LLM-based TCM formula models~\cite{ZhouArxiv2024-25,Chen2025TST} used LLM such as LLaMA~\cite{Dubey-Arxiv2024-23,Zhang2025TST} and ChatGLM~\cite{Zhipu-24} as the backbone model and constructed TCM formula instruction datasets to fine-tune LLM. This enables the models to provide TCM formula and explainable reasons. However, as the constructed datasets  lack fine-grained explainable reasons such as the composition of the formula's sovereign, minister, assistant, courier; efficacy; contraindications; tongue and pulse diagnosis, the fine-tuned LLM cannot provide this detailed information. Although Cui et al.~\cite{LiBIBM2021-6} constructed a Retrieval-Augmented Generation (RAG) framework based on a TCM knowledge graph to supplement TCM diagnosis, the built knowledge graph  focuses on disease diagnosis, prevention, and lifestyle management, which has little relevance to TCM formulas. Moreover, this RAG framework does not fine-tune LLMs, which prevents the LLMs from precisely integrating external retrieval information and causes suboptimal results.

To address the aforementioned issue, we propose~\RC, an AI-assisted tool designed to support both patients and clinicians. \sRC integrates GraphRAG with fine-tuning techniques for LLMs to solve the TCM formula task. More concretely,~\sRC uses GraphRAG to retrieve and synthesize structured TCM knowledge into concise summaries, while also constructing an enhanced instruction dataset to improve LLMs' ability to integrate retrieved information. 
We further provide novel theoretical proofs demonstrating  that \textbf{\emph {integrating GraphRAG with fine-tuning techniques}} can reduce generalization error and hallucination rates in TCM formula task. Experimental results on both collected and clinical datasets demonstrate that~\sRC achieves significant improvements over state-of-the-art models in all carefully designed quantitative metrics.
In summary, the contributions of this paper are as follows:

\noindent 1) We propose~\RC, the first model that integrates GraphRAG with fine-tuning techniques for LLMs to tackle the TCM formula task, and present novel theoretical proofs indicating this approach reduces generalization error and hallucination rates.

\noindent 2) Experimental results on both collected and clinical datasets demonstrate that~\sRC achieves significant improvements over state-of-the-art models.

\noindent 3) We open-source our model, implementation details in the appendix at https://huggingface.co/tczzx6/ZhiFangDanTai1.0. The dataset can be accessed at https://huggingface.co/datasets/tczzx6/ZhiFangDanTai1.0.

\section{Related Work}
\subsection{ Traditional Chinese Medicine Formula Models
}

Existing TCM formula models can be categorized into traditional TCM formula  models, deep learning-based TCM formula models, and LLM-based TCM formula models. Traditional TCM formula models employ association analysis~\cite{JingCNKI2024-1}, clustering~\cite{LiuTCM2017-2}, complex networks~\cite{NimaTCM2021-3}, and graph neural networks~\cite{ZhouPharmacol-10} to explore the associations between any two herbs in TCM formula. However, these methods cannot generate complete TCM formulas.  Deep learning-based TCM formula models aim to model the TCM diagnostic process of "wang (inspection), wen (listening and smelling), wen (inquiry), and qie (palpation)" through representation learning. 
Specifically, they first use convolutional neural networks~\cite{Zhang2024bioinformatics}, GCN~\cite{Wang2023,Lu2025Nature} or discriminative models~\cite{Wang2023bioinformation} to encode clinical indicators such as tongue coating characteristics, pulse patterns, and patient-reported symptoms; and then generate TCM formulas using established architectures including BERT~\cite{Chen2023}, Seq2seq models~\cite{LiuAI2022-21}, or GAN~\cite{Wang2023bioinformation}, with some approaches incorporating reinforcement learning~\cite{Wang2023} to optimize the recommendation quality.
Although these methods can provide complete TCM formula, they fail to give explainable reasons.
To solve  this problem, researchers propose LLM-based TCM formula models. One approach is to directly input symptoms into the LLM and request it to provide TCM formulas and explainable reasons~\cite{Kimi-27,Deepseek-28}.
However, due to the insufficient training data related to TCM formula in LLMs, these models tend to generate hallucinations and fail to provide convincing TCM formulas and reasons. Another approach is to construct TCM formula  instruction datasets to fine-tune the LLMs, enabling them to explain the rationale behind TCM formulas~\cite{ZhouArxiv2024-25,HaoJBU2024-30}. However, this method still cannot provide fine-grained information. Moreover, constructing TCM formula  instruction datasets requires domain expertise, which is labor-intensive and resource-consuming.

\subsection{ Retrieval-Augmented Generation}
The RAG framework was developed to mitigate key limitations of LLMs, including their inability to incorporate real-time knowledge updates, suboptimal performance in knowledge-intensive tasks, and lack of interpretability in question answering. RAG consists of three core stages: retrieval, augmentation, and generation~\cite{LewisNIPS2020-38}.
In the retrieval phase, researchers retrieve documents relevant to the query using Web APIs (e.g., Google Search API), segment these documents into smaller chunks, and encode them. Then they select top-$k$ documents most relevant to the query as the retrieved results.
In the augmentation phase, researchers concatenate the retrieved document segments with the original task instructions and inputs, which are then fed into the LLM.
In the generation phase, the LLM scans the entire input and generates responses based on its internal knowledge and the content of the retrieved documents.
However, RAG struggles to summarize or generalize effectively when relevant source content is unavailable~\cite{Edge-Arxiv-26}.

\subsection{ Graph-based Retrieval-Augmented Generation}
\label{subsec:GraphRAG}
To overcome the limitation of RAG in summarizing and generalizing content within text blocks, researchers have proposed the GraphRAG~\cite{Edge-Arxiv-26}. Specifically,  GraphRAG first extracts entities from text blocks to construct a knowledge graph and identifies fine-grained communities within the graph. Then, GraphRAG searches for answers to the query within these local communities and aggregates the partial answers to achieve summarization and generalization of the content within the text blocks. Currently, GraphRAG has been widely applied in NLP~\cite{YanArxiv2024-46}, AI agents~\cite{AnokhinArxiv2024-42} and recommender system~\cite{JieTongxin-49}. However, its application in the field of TCM formulas is still in the early stages of development. Moreover, rigorous theoretical proofs regarding the integration of GraphRAG with fine-tuned LLMs remains an unexplored area.

\section{\sRC Model  }

\begin{figure}[t]
	\centering
	\includegraphics[width= 0.5 \textwidth]{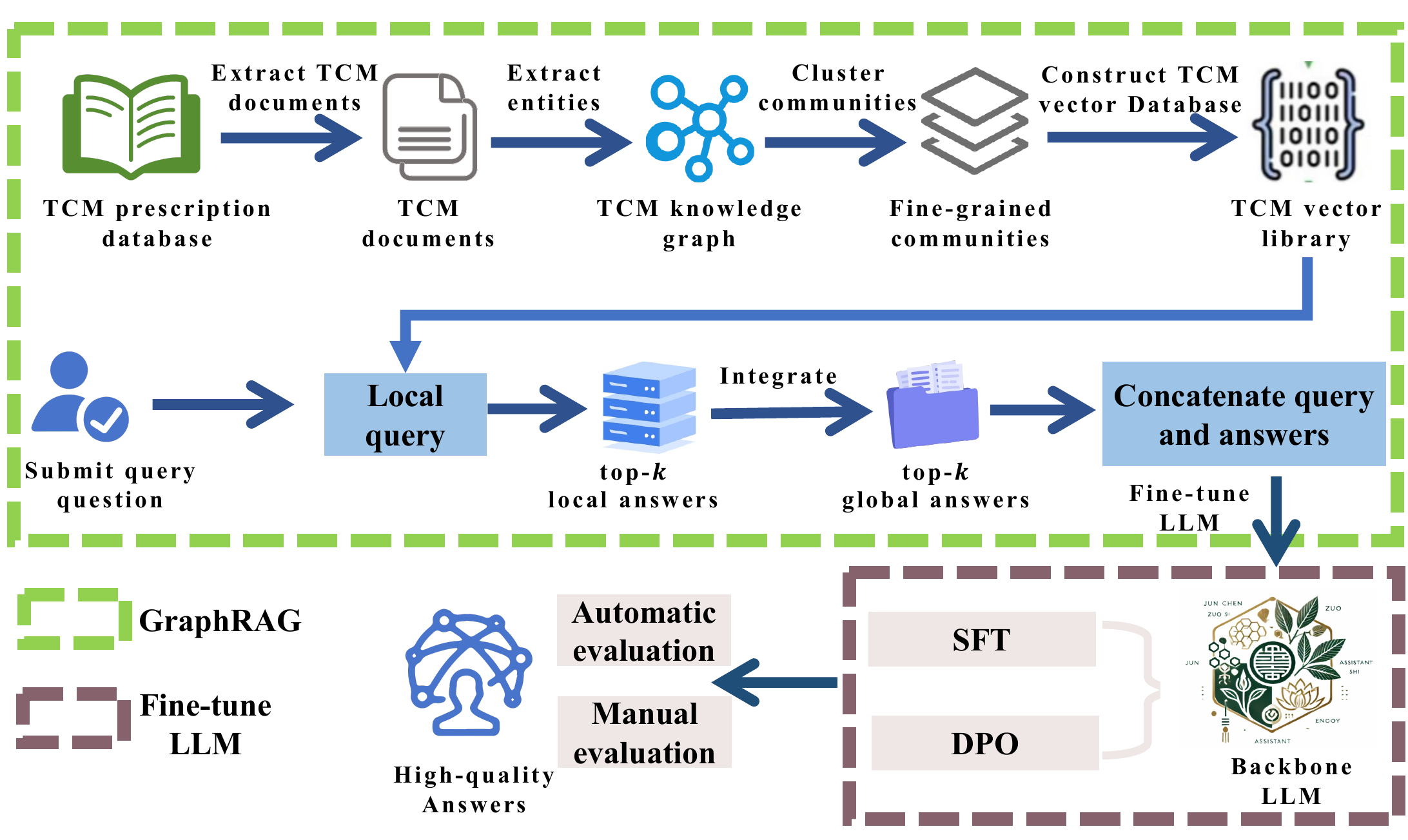}
	\caption{\label{fig:framework} The workflow of~\RC.~\sRC integrates GraphRAG with fine-tuning techniques for LLMs to solve the TCM formula task.
	}
\end{figure}

This section introduces the proposed model,~\RC, an AI-assisted tool designed to support both patients and clinicians. \sRC integrates GraphRAG with fine-tuning techniques for LLM to solve the TCM formula task.
Specifically, on one hand,~\sRC uses GraphRAG to retrieve fine-grained information and synthesize it into a concise and coherent summary in an efficient manner.
On the other hand,~\sRC constructs a TCM formula instruction dataset based on GraphRAG-generated results to perform Supervised Fine-tuning (SFT) and Direct Preference Optimization (DPO) on the LLM~\cite{JieTongxin-49}.
Fig.~\ref{fig:framework} illustrates the workflow of~\RC.
In the following sections, we first define the TCM formula problem and present six solutions from a final examination perspective, explaining how combining GraphRAG with fine-tuned LLMs improves TCM formula tasks. Next, we detail the GraphRAG pipeline, instruction fine-tuning, and model training/inference processes. Finally, we provide theoretical proofs showing that integrating GraphRAG with fine-tuning (SFT+DPO) reduces generalization error and hallucinations in TCM formula analysis.

\subsection{Problem Definition}
\label{subsec:problem_definition}
Given a description of symptoms $x$ and reference documents $\boldsymbol{\mathit{D}}$,  a model $\pi$ is required to generate a  reasonable TCM formula  and corresponding explanations $y$. The   explanations should include detailed information,  such as the composition of the formula's sovereign, minister, assistant, courier; efficacy; contraindications; tongue and pulse diagnosis.
Because LLMs have been proven to perform well in question answering tasks~\cite{Kimi-27}, we  initialize the model $\pi$ with an LLM. In order to answer 
why~\sRC integrates GraphRAG with fine-tuning technique for LLMs (cf. Section~\ref{sec:introduction}), we illustrate six solutions——plug and play; fine-tuning LLMs (FT); RAG; GraphRAG; RAG+FT; GraphRAG+FT, from  the perspective of a final examination as follows:

\begin{table*}
	\centering
	\caption{Two examples of the original TCM formula dataset. Chinese originals are translated into English.}
	\label{tb:original_tcm}
	\small
	\begin{tabular}{p{2.6cm}|p{2.2cm}|p{2cm}|p{4cm}|p{4.2cm}p{0.001cm}}
		\toprule
		\centering Disease &\centering Formula&\centering Processing Methods&\centering Formula Composition&  \centering Usage Method & \\ 
		\midrule 
		Chronic Intestinal Wind with Bleeding &  Halloysite Decoction   &Honey-frying, Peel-removing, Wine-frying&Halloysite Coptis, Magnolia Bark, Licorice Root, Chinese Angelica&Mix all herbs in the specified proportions, decoct with water, and consume the resulting liquid. & \\ \midrule
		Postpartum Bloody Dysentery 
		Mixed Red-White Bleeding
		& Halloysitum Rubrum Decoction		&Coarsely ground ingredients&Halloysite rubrum, Coptis root, Sanguisorba root, Prepared licorice, Magnolia bark, Dry ginger, Angelica root&Take 3 qianbi per dose. Boil in 1 cup water with 3-inch cut Chinese chives. Decoct until 70\% liquid remains. & \\
		
		\bottomrule
	\end{tabular}
\end{table*}

\noindent \textbf{1) Plug-and-play~\cite{Thirunavukarasu2023Nature}}: Asking the LLM to generate an answer directly from symptoms is like a closed-book exam—the model relies only on its internal knowledge without external references.

\noindent \textbf{2) FT~\cite{Bao2024Recsys}}: Create a TCM formula instruction dataset and fine-tune the LLM to generate symptom-based answers. This resembles reviewing material before a closed-book exam.

\noindent \textbf{3) RAG~\cite{XiongACL2024}}: Require the LLM generate answers using symptoms and unstructured reference documents. This is like an open-book exam with unsorted reference materials.

\noindent \textbf{4) GraphRAG~\cite{Wu2024arxiv}}: Ask the LLM to generates answers using symptoms and reference documents partitioned into $n$ subtopics via community detection. This mirrors an open-book exam with systematically organized materials.

\noindent \textbf{5) RAG+FT~\cite{Giuffre2025DLD}}: Fine-tune the LLM using a dataset where it generates answers from symptoms and unstructured reference documents. This process resembles a practice open-book exam before the actual one—both allow referencing mixed, unorganized materials.

\noindent \textbf{6) GraphRAG+FT}: Fine-tune the LLM using an instruction dataset where it generates answers from symptoms and reference documents partitioned into $n$ subtopics via community detection. This mimics a practice open-book exam with categorized materials before the actual exam.

Existing approaches (Solutions 1) - 5)) face limitations in addressing TCM formula tasks. Solutions 1) - 2) lack reference documents, relying solely on the model's internal knowledge, which proves inadequate for domain-specific challenges. While Solutions 3) - 5) incorporate reference documents, they either lack fine-grained division (hindering precise retrieval) or omit LLM fine-tuning (limiting information integration). In contrast, Solution 6) combines GraphRAG's summarization/generalization capabilities with fine-tuned LLMs, enabling precise adaptation to TCM formula tasks. Therefore,~\sRC implements Solution 6), integrating GraphRAG with LLM fine-tuning for optimal performance.

\subsection{ GraphRAG}
Current TCM formula datasets provide descriptions but lack detailed explanations. To generate comprehensive explanations, it is necessary to augment these datasets with fine-grained information from external sources. As discussed in Section~\ref{subsec:GraphRAG}, both RAG and GraphRAG can retrieve useful knowledge for improving LLM-generated responses. However, GraphRAG outperforms RAG in summarizing and generalizing text-block content. To this end, we utilize GraphRAG to solve the TCM formula task.
Following~\cite{Edge-Arxiv-26}, we divide GraphRAG into retrieval and generation phases. In retrieval phrase (Section~\ref{subsec:collect_dataset} - Section~\ref{subsec:community_summaries}), we build a knowledge graph $\mathcal{G}$ from reference documents $\boldsymbol{\mathit{D}}$ and derive $n$ community summaries by clustering related entities. In  generation phrase (Section~\ref{subsec:rag}), given symptom descriptions $x$, we adopt a map-reduce approach~\cite{DeanACM-52}: the map phase retrieves answers from each community summary, while the reduce phase summarize these fine-grained results. The combined output—along with the original question—is then fed into the LLM to further perform SFT and DPO phrases. The details of GraphRAG are as follows:

\subsubsection{Collect Original Dataset$\rightarrow$Create Source Documents }
\label{subsec:collect_dataset}
We collect 80,000 pieces of TCM formula data from the Chinese Formula Database\footnote{https://www.ncmi.cn/phda/dataDetails}, Ancient Chinese Formula Database\footnote{https://www.ncmi.cn/phda/dataDetails}, the Catalogue of Ancient Classic Formulas published by the National Administration of Traditional Chinese Medicine\footnote{https://www.gov.cn/zhengce/zhengceku/2018-12/31/content\_5429153.html}, and the National Medical Products Administration\footnote{https://www.nmpa.gov.cn/xxgk/fgwj/gzwj/gzwjyp/20230531161831118.html}. The collected dataset includes disease names, formula components, preparation methods, and usage instructions from classical texts such as Huangdi Neijing, Bencao Gangmu, and Shanghan Zabing Lun (see Table~\ref{tb:original_tcm} for examples). Since the dataset lacks fine-grained information such as the composition of the formula's sovereign, minister, assistant, courier; efficacy; contraindications; tongue and pulse diagnosis,  it cannot be directly used to fine-tune LLMs for generating TCM formulas and corresponding explanations.

To address the absence of fine-grained information in the original TCM dataset, we employ DeepSeek~\cite{Deepseek-28} to extract detailed information for each prescription using its name and ingredients as input.
Leveraging LLMs' in-context learning capability~\cite{OuyangNIPS2022-50}, we first annotate 50 samples as reference examples, then utilize the model to verify and refine the remaining data through redundant cleaning and formula interpretation. The resulting documents  $\boldsymbol{\mathit{D}}$  contain seven key elements for each symptom $x$,
i.e., \textit{Disease}, \textit{Recommended Formulas}, \textit{Herbal Ingredients},  \textit{Applicable Symptoms and Population}, \textit{Pulse and Tongue Diagnosis}, \textit{Contraindications} and \textit{Preparation Methods}.

\subsubsection{ Source Documents $\rightarrow$ Text Chunks }
 We split $\boldsymbol{\mathit{D}}$ into several text chunks. 
Selecting the chunk size is a trade-off issue, as shorter text chunks suffer from the high recall problem, while longer text chunks introduce redundant information and increases processing time for large models~\cite{Edge-Arxiv-26}. Through empirical findings, we set the chunk size to 512 tokens. 

\subsubsection{  Text Chunks $\rightarrow$  Entities \& Relations }
We prompt the LLM  to extract instances of entities and the relationships
between entities from a given chunk, since recent studies by Wei~\cite{WeiArxiv2023} and Lopez et al.~\cite{Lopez2025Digital} have demonstrated that LLMs outperform state-of-the-art deep learning models in entity extraction tasks. Based on this finding, we employ an LLM (e.g., DeepSeek-7B) for entity-relation extraction and subsequent relationship categorization. We define two types of relationships between entities: intra-category relationships occur when both entities belong to the same sub-community, while inter-category relationships exist when entities originate from different sub-communities.


\subsubsection{   Entities \& Relations $\rightarrow$    Knowledge Graph }
\begin{figure}[t]
	\centering
	\includegraphics[width=8.0cm,height=6.0cm]{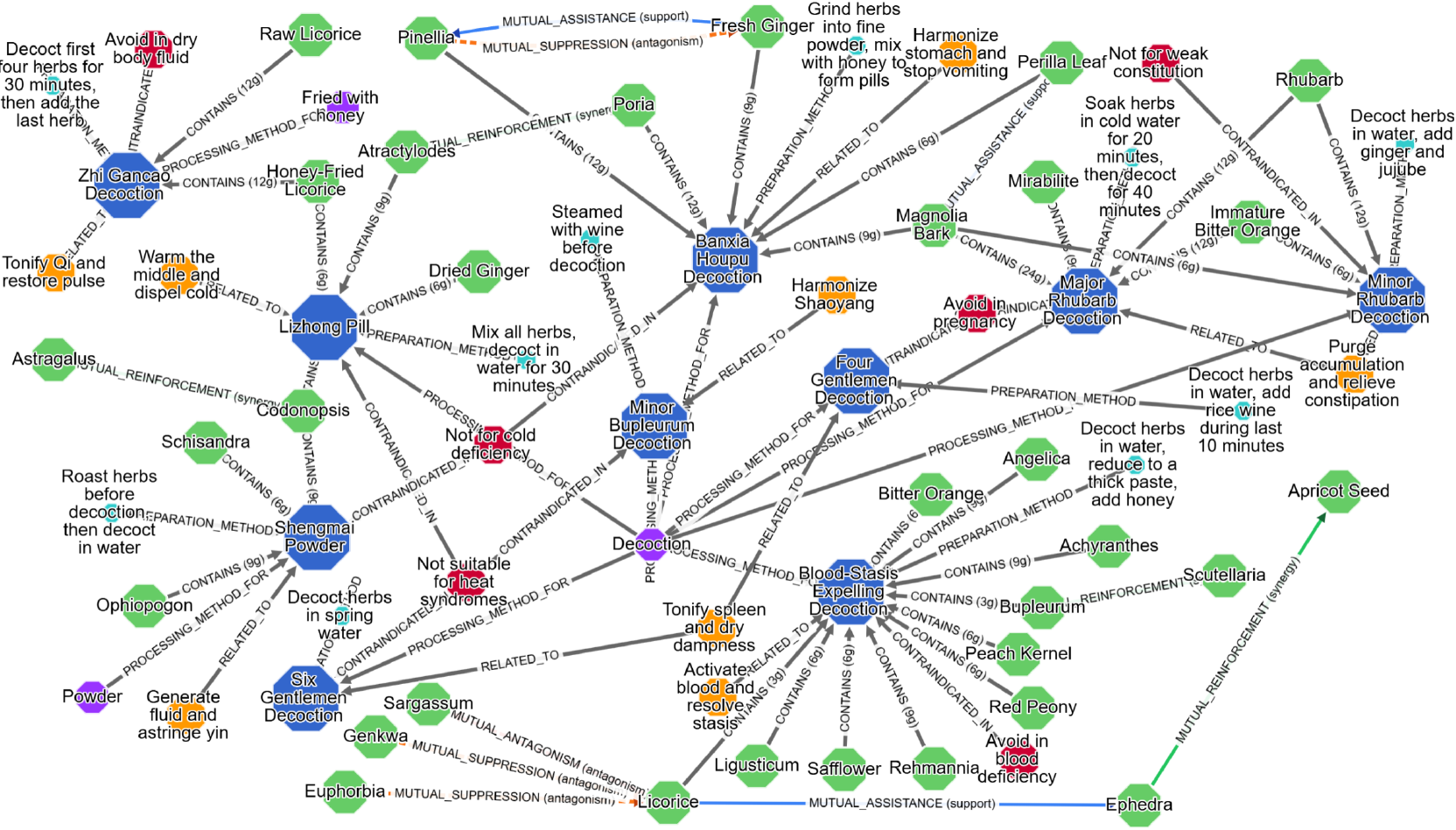}
	\caption{\label{fig:kg} Constructed Knowledge Graph (Partial). We translate the knowledge graph (KG) into English for clarity.
	}
\end{figure}

Base on the extracted entities and relations, we construct  knowledge graph $\mathcal{G}$ using neo4j (https://neo4j.com/), which enables quick location of \textit{fine-grained information} and improves search accuracy during retrieval.
We define two types of relationships between entities: inter-category relationships occur when both entities belong to the same sub-community $\vec{C}_k$, while intra-category relationships exist when entities originate from different sub-communities
Fig.~\ref{fig:kg} shows a portion of  the constructed knowledge graph $\mathcal{G}$, whereas edges connecting nodes of the same color indicate intra-category relationships, while edges between differently colored nodes denote inter-category relationships.


\subsubsection {  Knowledge Graph $\rightarrow$ Graph Communities  $\rightarrow$ Community Summaries }
\label{subsec:community_summaries}
Based on the knowledge graph $\mathcal{G}$,  we identify diverse communities to analyze TCM formulas from multiple perspectives.  We apply the Leiden community algorithm~\cite{Traag2019-51} in a hierarchical manner to detect sub-communities until no further splitting is possible.
The resulting communities are divided into seven categories, matching our  \textit{fine-grained information} taxonomy, i.e., \textit{Diseases}, \textit{Recommended Formulas}, \textit{Herbal Ingredients}, \textit{Applicable Symptoms and Population}, \textit{Pulse and Tongue Diagnosis}, \textit{ Contraindications}, and \textit{Preparation Methods}. Formally, we use $ C = \{C_1, \ldots, C_7\} $ to represent the detected communities.
Each community $ C_i $ contains a set of entities $ \{e_1, \ldots, e_{|C_i|}\} $ and a community description $ s_{|C_i|} $. Table~\ref{tb:community}  presents the basic information of the extracted fine-grained communities from the document.
After community division, we encode them into a vector library $ \vec{C} = \{\vec{C_1}, \ldots, \vec{C_7}\} $. For implementation, we use LlamaIndex Library\footnote{https://docs.llamaindex.ai/en/stable/}  for retrieval, with mxbai-embed-large\footnote{https://inscode-doc.inscode.cc/} as the default encoder. Other encoders like OpenAI's text-embedding-ada-002 can also be used.

\begin{table*}
	\centering
	\caption{Fine-Grained Community Information.}
	\label{tb:community}
	\small
	\begin{tabular}{p{3.6cm}|p{3.2cm}|p{3.6cm}|p{4.5cm}p{0.001cm}}
		\toprule
		\centering Community Name &\centering Entity Examples&\centering Community Description&\centering Example&   \\ 
		\midrule 
	 	Disease &  Intestinal Wind Bleeding, Wind-Cold Common Cold, Wind-Heat Common Cold... &Description of disease patterns:&Intestinal Wind Bleeding \\ \midrule
		Recommended Formulas
		& Intestinal Wind Bleeding Pill, Cinnamon Twig Decoction, Minor Bluegreen Dragon...		&Effective herbal formulas for specific diseases&Intestinal Wind Bleeding Pill&  \\ \midrule
		Herbal Components & Dried persimmon cake, Black plum, Oil-jar indocalamus leaf, Sophora flower, Medicinal gallnut, Bitter orange...&Composition of monarch, minister, assistant, courier herbs and their synergies&[Herbal Components] Monarch drug: Dried persimmon cake (2 liang): Astringes intestines, stops bleeding. Mainly treats intestinal wind bleeding.:... \\ \midrule
		Applicable Symptoms  and Population & Bloody stools, abdominal pain, bloating, shortness of breath... & Target patient profiles & [Applicable Symptoms] Indicated for intestinal wind bleeding patients presenting bloody stools, abdominal pain, bloating,... \\ \midrule
		Pulse and Tongue Diagnosis&  Thin white coating, yellow greasy coating, fissured tongue... & Diagnostic tongue and pulse characteristics & [Pulse-Tongue Signs] Typical pulses: thready-weak or slippery, indicating Qi-blood deficiency or damp-heat. Tongue: pale-red ... \\ \midrule
		Contraindications & Yin deficiency with fire excess, spleen-stomach deficiency cold... & Usage restrictions & [Contraindications] Avoid in pregnancy, Yin deficiency with fire... \\ \midrule
		Preparation Methods & Honey-frying, Peel-removing, Wine-frying... & Special preparation techniques & [Processing Methods] Required steps: Char persimmon cakes/plums/indocalamus leaves, ... \\
		
		\bottomrule
	\end{tabular}
\end{table*}

\subsubsection{ Community Summaries  $\rightarrow$  Community Answers $\rightarrow$ Global Answer}\label{subsec:fine-tuning}
\label{subsec:rag}
Given a symptom description $ x $ and the fine-grained community vector library $ \vec{C} = \{\vec{C_1}, \ldots, \vec{C_7}\} $, we employ query expansion techniques~\cite{YanArxiv2024-46} to expand the symptom description $x$, resulting in the expanded query description $\{x, x'\}$. We then conduct local information retrieval for each community $ C_i $ to obtain local answers $\{A_1, \ldots, A_7\}$. 
Formally, the process is as follows: the content of the $ i $-th local information retrieval is given by $ p(A_i | \{x, x'\}) \propto \exp(\textbf{E}(A_i)\textbf{E}(x \| x')) $, where $\textbf{E}$ is the encoder, and $ \| $ denotes the concatenation operation. Without loss of generality, we also select mxbai-embed-large as the encoder. Other encoders, including the text-embedding-ada-002 encoder from the ChatGPT series, may also be considered as viable options.
Following local information retrieval, we employ the LLM to synthesize local responses into a comprehensive global answer $c$. Formally, this can be expressed as  $c = \pi(x, x', A_1, \ldots, A_7) $. Equivalently, this operation can be represented as $c = \pi(x, x', \mathcal{G}_x^{*})$, where $\mathcal{G}_x^{*}$ is the relevant sub-graph of $\mathcal{G}$ that matches the local answer $ \{ A_1, \ldots, A_7 \}$. 

Notably, given a symptom $x$, the above retrieval operations can only yield one set of local retrieval answers $\{A_1, \ldots, A_7\}$ and one global retrieval answer $c$. To enrich the retrieval results, we employ the Beam Search~\cite{MeisterTACL2020-47} method during the local/global information retrieval process to obtain top-$k$ local retrieval answers and top-$k$ global retrieval answers. We discuss the impact of the top-$k$ setting in Section~\ref{subsec:topk}.

\hide{\footnote{We refer top-$k$ as \textit{retrieval item count} in the subsequent discussions.}}

\subsection{ Fine-tuning Large Language Model}

Directly using GraphRAG can enhance the ability of LLMs to provide \textit{fine-grained information} of the TCM formula task. However, as mentioned in Section~\ref{subsec:problem_definition}, combining GraphRAG and fine-tuning technique can further improve the ability of LLMs to integrate external retrieval information. To this end, we further fine-tune LLM based on the fine-grained instruction dataset. As suggested by Zhou et al.~\cite{Zhou-32}, fine-tuning LLMs consists of three continuous steps, i.e., Generative Pre-training (GPT), SFT, and DPO~\cite{Rafailov-33}. However, GPT is  not required when  the base LLM already demonstrates sufficient capability and performance. Thus, we only adopt SFT and DPO to fine-tune the LLM. Without loss of generality, we select LLaMA3.2-7B~\cite{Dubey-Arxiv2024-23} as the base LLM model $\pi$. The details are as follows:
\subsubsection{ SFT}
Given the answer generated by GraphRAG and the corresponding ground-truth answer, we conduct SFT to optimize the parameters of the base LLM $\pi$. This optimization aims to enhance the accuracy of both GraphRAG and the fine-tuned LLM. The loss function is:
\begin{equation}
\label{sft}
\mathcal{L}_{\text{GraphRAG+SFT}} = -\mathbb{E}_{(x,c,y)\sim D}\log P_{\theta}(y|x,c),
\end{equation}
\noindent where $(x,c,y)\sim D$ represents sampling an data piece from the
instruction dataset  $ D = \{ Instruction, Input, Output \} $. The $Instruction$ content is defined as "Recommend a TCM formula and provide detailed explanations based on the  symptoms".  The $Input$  content includes the symptom description $x$ and the global retrieval answer $c$. The $Output$ content $y$ is the ground-truth answers, which are obtained through human annotation and in-context learning with a LLM.

\subsubsection{ DPO}
Unlike Reinforcement Learning from Human Feedback (RLHF)~\cite{OuyangNIPS2022-50}, DPO eliminates the need for training a separate reward model~\cite{Rafailov-33}. Formally, given a description of symptoms $x$ and generated global answer $c$,  we require the LLM $ \pi_\text{ref} $ generating two answers and score them according to quality of the answers. We denote the answer with  higher score  as $ y_w $, and the one with lower score  as $ y_l $, i.e., $(y_w, y_l)=\pi_{\text{ref}}(x, c)$. Then, we minimize the DPO loss  function $\mathcal{L}_\text{DPO}(\pi_{\theta}, \pi_{\text{ref}})$:

\begin{equation}
\label{dpo}
-\mathbb{E}_{p \sim D}[ \log \sigma ( \beta \log \frac{\pi_{\theta}(y_w|x)}{\pi_{\text{ref}}(y_w|x)} - 
 \log \frac{\pi_{\theta}(y_l|x)}{\pi_{\text{ref}}(y_l|x)} ) ],
\end{equation}

\noindent where $p=(x,c, y_w, y_l)$ denotes data sampling process, $ \sigma$ is the  sigmoid function, $ \pi_{\theta}$ is the LLM  to be optimized and is initialized as $ \pi_{\text{ref}} $.

\subsection{Model Training}
We implement the GraphRAG module using the LlamaIndex library and integrate it with LlamaFactory~\footnote{https://github.com/hiyouga/LLaMA-Factory} for SFT and DPO.
Specifically, we first obtain ground-truth answers through human annotation and in-context learning with a LLM. Then, we leverage LlamaIndex to prompt the LLM to generate answers, and training LLM using Eq~\eqref{sft} and Eq~\eqref{dpo}.

\subsection{Model Inference}
\label{subsec:model_inference}
During the model inference phase, given symptoms $x$, we first employ the query expansion technique to obtain the expanded query description $\{x, x'\}$, and then retrieve local answers  $\{A_1, \ldots, A_7\}$ based on the fine-grained communities, followed by employing the LLM $\pi_{\theta}$ to obtain the final global retrieval answer $c$. Finally,  based on $\{x, x'\}$ and $c$, we ask the LLM  $\pi_{\theta}$ to generate the  TCM formula and corresponding explanations $y$.
In practice, we  use vLLM and deploy~\sRC using the WebUI framework that comes with LLaMA Factory to accelerate the model's inference process.

\subsection{Theoretical proof}
We present rigorous theoretical proofs showing that combining GraphRAG with fine-tuning techniques (SFT+DPO) reduces generalization error and hallucination rates in TCM formula tasks.

\noindent \subsubsection{Proposition 1} \textbf{GraphRAG+SFT reduce  generalization error
\  \ Let $\boldsymbol{I}(y; c | x)$ be the conditional mutual information between GraphRAG-generated content $c(x, G)$ and target answer $y$. When  $\boldsymbol{I}(y; c | x) \geq \gamma$, combining GraphRAG with SFT yields a generalization error $\mathcal{E}(\theta_{\text{GraphRAG+SFT}})$ bounded by}

\begin{equation}
\mathcal{E}(\theta_{\text{GraphRAG+SFT}}) \leq \mathcal{E}(\theta_{\text{SFT}}) - \frac{\gamma}{\beta},
\end{equation}

\noindent where $\gamma$ represents the minimum mutual information threshold, $\beta$ is a positive constant.

\noindent  \textbf{Proof:} \ \ The conditional mutual information  $\boldsymbol{I}(y; c | x)$ measures the contribution of context $c$ to prediction $y$, which can be expressed as:
\begin{equation}
\nonumber
\boldsymbol{I}(y; c | x) = \mathbb{E}_{(x,y,c)} \left[ \log \frac{P(y \mid x, c)}{P(y \mid x)} \right].
\end{equation}

\noindent  According to Eq.~\eqref{sft}, we have the generalization error of GraphRAG+SFT:
\begin{equation}
\nonumber
\mathcal{E}(\theta_{\text{GraphRAG+SFT}}) = \mathbb{E}\left[ -\log P_\theta (y \mid x, c) \right].
\end{equation}

\noindent  Transform formula $\log P (y \mid x, c)$, we have:
\begin{equation}
\nonumber
\begin{aligned}
\log P(y \mid x, c) &= \log \left[ P(y \mid x) \cdot \frac{P(y \mid x, c)}{P(y \mid x)} \right] \\
&= \log P(y \mid x) + \log \frac{P(y \mid x, c)}{P(y \mid x)}.
\end{aligned}
\end{equation}

\noindent  By taking the negative expectation of both sides, we obtain:
\begin{equation}
\nonumber
\mathbb{E}[-\log P(y \mid x, c)] = \mathbb{E}[-\log P(y \mid x)] - \mathbb{E}\left[\log \frac{P(y \mid x, c)}{P(y \mid x)}\right].
\end{equation}

\noindent  Since $\mathbb{E}\left[\log \frac{P(y \mid x, c)}{P(y \mid x)}\right] = \boldsymbol{I}(y; c | x) $, $\boldsymbol{I}(y; c | x)  \geq \gamma$ and $\mathbb{E}[-\log P(y \mid x) = \mathcal{E}(\theta_{\text{SFT}})$, we have:
\begin{equation}
\nonumber
\mathcal{E}(\theta_{\text{GraphRAG+SFT}}) = \mathcal{E}(\theta_{\text{SFT}}) - I(y; c \mid x) \leq \mathcal{E}(\theta_{\text{SFT}}) - \gamma.
\end{equation}

\noindent  However, in practice, we use backpropagation to optimize LLM, thus the optimization of $\boldsymbol{I}(y; c | x)$ needs to be scaled by $\beta$, i.e.,
\begin{equation}
\nonumber
\Delta \mathcal{E} \geq \frac{\boldsymbol{I}(y; c | x)}{\beta} \geq \frac{\gamma}{\beta},
\end{equation}

\noindent  where $\Delta \mathcal{E} = \mathcal{E}(\theta_{\text{SFT}}) - \mathcal{E}(\theta_{\text{GraphRAG+SFT}})$, the proof is as follows:

\noindent  Assume the loss function $\mathcal{L}_{\text{GraphRAG+SFT}}$ is $\beta$-smooth, i.e., its gradient satisfies the Lipschitz condition:

\begin{equation}
\nonumber
\|\nabla_\theta \mathcal{E}(\theta) - \nabla_\theta \mathcal{E}(\theta')\| \leq \beta\|\theta - \theta'\|,
\end{equation}

\noindent  where $\theta$ and $\theta'$ represent model parameters at any two distinct training epochs during the optimization process. The Lipschitz condition is equivalent to the absolute value of the eigenvalues of the Hessian matrix not exceeding $\beta$, that is:
\begin{equation}
\nonumber
\|\nabla^2 \mathcal{E}(\theta)\| \leq \beta
\end{equation}

\noindent  According to the Taylor Expansion, we have:
\begin{equation}
\nonumber
\begin{aligned}
\mathcal{E}(\theta_{t+1}) &=\mathcal{E}(\theta_t) + \nabla \mathcal{E}(\theta_t)^\top (\theta_{t+1} - \theta_t) \\ &+ \frac{1}{2} (\theta_{t+1} - \theta_t)^\top \nabla^2 \mathcal{E}(\xi) (\theta_{t+1} - \theta_t) \\ &\leq \mathcal{E}(\theta_t) + \nabla \mathcal{E}(\theta_t)^\top (\theta_{t+1} - \theta_t)  + \frac{\beta}{2}\|\theta_{t+1} - \theta_t\|^2.
\end{aligned}
\end{equation}

\noindent  where $\xi$ is any data point between $\theta_{t+1}$ and $\theta_t$, $\nabla^2 \mathcal{E}(\xi)$ is the Hessian matrix.

\noindent  The update rule for gradient descent is:
\begin{equation}
\nonumber
\theta_{t+1} = \theta_t - \eta \nabla \mathcal{E}(\theta_t).
\end{equation}

\noindent  Substitute it into Taylor Expansion, the first-order term is:
\begin{equation}
\nonumber
\nabla \mathcal{E}(\theta_t)^\top (\theta_{t+1} - \theta_t) = \nabla \mathcal{E}(\theta_t)^\top (-\eta \nabla \mathcal{E}(\theta_t)) = -\eta \|\nabla \mathcal{E}(\theta_t)\|^2.
\end{equation}

\noindent  The second-order term is:
\begin{equation}
\nonumber
\frac{\beta}{2}\|\theta_{t+1} - \theta_t\|^2 = \frac{\beta}{2} \|-\eta \nabla \mathcal{E}(\theta_t)\|^2 = \frac{\beta \eta^2}{2} \|\nabla \mathcal{E}(\theta_t)\|^2.
\end{equation}

\noindent  Merge the first-order and the second-order term, we have:
\begin{equation}
\nonumber
\mathcal{E}(\theta_{t+1}) \leq \mathcal{E}(\theta_t) - \eta \|\nabla \mathcal{E}(\theta_t)\|^2 + \frac{\beta \eta^2}{2} \|\nabla \mathcal{E}(\theta_t)\|^2.
\end{equation}

\noindent  Merge term $\|\nabla \mathcal{E}(\theta_t)\|^2$:
\begin{equation}
\nonumber
\mathcal{E}(\theta_{t+1}) \leq \mathcal{E}(\theta_t) - \left(\eta - \frac{\beta \eta^2}{2}\right) \|\nabla \mathcal{E}(\theta_t)\|^2.
\end{equation}

\noindent  To maximize the error reduction, $\eta$ should be chosen to maximize $\eta - \frac{\beta \eta^2}{2}$.

\noindent  Take the derivative of $\eta$ and set the derivative to zero, we have:
\begin{equation}
\nonumber
\eta = \frac{1}{\beta}.
\end{equation}

\noindent  Thus, the inequality becomes:
\begin{equation}
\nonumber
\mathcal{E}(\theta_{t+1}) \leq \mathcal{E}(\theta_t) - \frac{\|\nabla \mathcal{E}(\theta_t)\|^2}{2\beta}.
\end{equation}

\noindent   Since the information gain $\boldsymbol{I}(y; c | x)$ establishes a lower bound on the expected gradient alignment, ensuring that each optimization step reduces the error by at least:
\begin{equation}
\nonumber
\|\nabla \mathcal{E}(\theta_t)\|^2 \geq \boldsymbol{I}(y; c | x)^2.
\end{equation}

\noindent  Therefore, the generalization error is guaranteed to decrease by at least:
\begin{equation}
\nonumber
\mathcal{E}(\theta_{t+1}) \leq \mathcal{E}(\theta_t) - \frac{ \boldsymbol{I}(y; c | x)^2}{2\beta}.
\end{equation}

\noindent  Consider $ \boldsymbol{I}(y;c|x) \geq \gamma$, by substituting $\theta_{\text{GraphRAG+SFT}}$ and $\theta_{\text{SFT}}$ for $\theta_{t+1}$ and $\theta_t$ respectively in the error bound, we obtain:
\begin{equation}
\nonumber
\mathcal{E}(\theta_{\text{GraphRAG+SFT}}) \leq \mathcal{E}(\theta_{\text{SFT}}) - \frac{\boldsymbol{I}(y;c|x)^2}{2\beta} \leq \mathcal{E}(\theta_{\text{SFT}}) - \frac{\gamma^2}{2\beta}.
\end{equation}

\noindent  After integrating over $T$ optimization steps, the total generalization error reduction satisfies:
\begin{equation}
\nonumber
\mathcal{E}(\theta_{\text{GraphRAG+SFT}}) \leq \mathcal{E}(\theta_{\text{SFT}}) - \frac{\gamma}{\beta}.
\end{equation}

\noindent  Since the integration of GraphRAG enhances the information gain $\gamma$, which directly reduces the upper bound on the generalization error $\mathcal{E}(\theta_{\text{GraphRAG+SFT}})$. This demonstrates that combining GraphRAG with SFT yields provably better generalization compared to SFT alone.

\noindent   \subsubsection{Proposition 2} \textbf{Incorporating DPO can further reduce the generalization error} \ \ Building upon the combined GraphRAG and SFT approach, the incorporation of DPO yields the following generalization error bound for preference-aligned data:
\begin{equation}
\mathcal{E}(\theta_{\text{DPO}}) \leq \mathcal{E}(\theta_{\text{SFT}}) - \frac{\mathbb{E}[\Delta]}{\beta},
\end{equation}
where $\Delta = \log \frac{P_{\text{ref}}(y_w|x,c)}{P_{\text{ref}}(y_l|x,c)}$ represents preference strength.

\noindent \textbf{Proof:}
The optimal solution for DPO ensures:
\begin{equation}
\nonumber
P_\theta(y|x,c) = \frac{1}{Z(x,c)} P_{\text{ref}}(y|x,c) e^{\beta^{-1} r(x,y)},
\end{equation}
where $r(x,y)$ is implicit reward. For preference pairs $(y_w,y_l)$, we have:
\begin{equation}
\nonumber
\frac{P_\theta(y_w|x,c)}{P_\theta(y_l|x,c)} = \frac{P_{\text{ref}}(y_w|x,c)}{P_{\text{ref}}(y_l|x,c)} e^{\beta^{-1} (r(x,y_w) - r(x,y_l))}.
\end{equation}

\noindent  For a given preference pair $(y_w,y_l)$, the DPO preference probability model is expressed as:
\begin{equation}
\nonumber
P(y_w \succ y_l|x) = \sigma\left(\beta \log \frac{P_\theta(y_w|x,c)}{P_{\text{ref}}(y_w|x,c)} - \beta \log \frac{P_\theta(y_l|x,c)}{P_{\text{ref}}(y_l|x,c)}\right).
\end{equation}

\noindent  The optimization objective is to maximize $P(y_w \succ y_l|x)$, i.e.:
\begin{equation}
\nonumber
\beta \log \frac{P_\theta(y_w|x,c)}{P_{\text{ref}}(y_w|x,c)} - \beta \log \frac{P_\theta(y_l|x,c)}{P_{\text{ref}}(y_l|x,c)} \geq \Delta.
\end{equation}

\noindent  By expanding the inequality, we obtain:
\begin{equation}
\nonumber
\log \frac{P_\theta(y_w|x,c)}{P_{\text{ref}}(y_w|x,c)} - \log \frac{P_\theta(y_l|x,c)}{P_{\text{ref}}(y_l|x,c)} \geq \beta^{-1} \Delta.
\end{equation}

\noindent  By applying the property of logarithmic subtraction:
\begin{equation}
\nonumber
\log\left(\frac{P_\theta(y_w|x,c)}{P_{\text{ref}}(y_w|x,c)} \cdot \frac{P_{\text{ref}}(y_l|x,c)}{P_\theta(y_l|x,c)}\right) \geq \beta^{-1} \Delta.
\end{equation}

\noindent  After taking the exponential:
\begin{equation}
\nonumber
\frac{P_\theta(y_w|x,c)}{P_{\text{ref}}(y_w|x,c)} \cdot \frac{P_{\text{ref}}(y_l|x,c)}{P_\theta(y_l|x,c)} \geq e^{\beta^{-1} \Delta}.
\end{equation}

\noindent  As DPO decreases the likelihood of $y_l$, we have:
\begin{equation}
\nonumber
\frac{P_{\text{ref}}(y_l|x,c)}{P_\theta(y_l|x,c)} \geq 1.
\end{equation}

\noindent  Thus:
\begin{equation}
\nonumber
\frac{P_\theta(y_w|x,c)}{P_{\text{ref}}(y_w|x,c)} \geq e^{\beta^{-1} \Delta}.
\end{equation}

\noindent  Therefore:
\begin{equation}
\nonumber
P_\theta(y_w|x,c) \geq P_{\text{ref}}(y_w|x,c) e^{\beta^{-1} \Delta}.
\end{equation}

\noindent  Take negatives of both sides and apply logarithms:

\begin{equation}
\label{eq:1}
\begin{aligned}
-\log P_\theta(y_w|x,c) &\leq -\log P_{\text{ref}}(y_w|x,c) - \beta^{-1}\Delta.
\end{aligned}
\end{equation}

\noindent  Similarly, since DPO  increases the likelihood of $y_w$, we have:
\begin{equation}
\nonumber
\frac{P_\theta(y_w|x,c)}{P_{\text{ref}}(y_w|x,c)} \geq 1, {\rm thus,}
\end{equation}

\begin{equation}
\nonumber
\frac{P_{\text{ref}}(y_l|x,c)}{P_\theta(y_l|x,c)} \leq e^{\beta^{-1}\Delta}, {\rm thus,}
\end{equation}

\begin{equation}
\nonumber
P_{\text{ref}}(y_l|x,c) \leq P_\theta(y_l|x,c) e^{\beta^{-1}\Delta}.
\end{equation}

\noindent Take negatives of both sides and then apply logarithms:
\begin{equation}
\label{eq:2}
\begin{aligned}
-\log P_{\text{ref}}(y_l|x,c) &\geq -\log P_\theta(y_l|x,c) - \beta^{-1}\Delta, {\rm thus,} \\
-\log P_\theta(y_l|x,c) &\leq -\log P_{\text{ref}}(y_l|x,c) + \beta^{-1}\Delta
\end{aligned}
\end{equation}

\noindent The generalization error of DPO primarily stems from distributional shifts in $y_w$ and $y_l$:

\begin{equation}
\nonumber
\begin{aligned}
\nonumber
\mathcal{E}(\theta_{\text{DPO}}) = \mathbb{E} [-\log P_\theta(y_w|x,c) \cdot \mathbb{I}(y=y_w) \\ - \log P_\theta(y_l|x,c) \cdot \mathbb{I}(y=y_l) ]
\end{aligned}
\end{equation}

\noindent Substituting Eq.~\eqref{eq:1} and~\eqref{eq:2}  into the above equation, we have:
\begin{align*}
\mathcal{E}(\theta_{\text{DPO}}) &\leq \mathbb{E}\left[\left(-\log P_{\text{ref}}(y_w|x,c) - \beta^{-1}\Delta\right) \cdot \mathbb{I}(y=y_w) \right. \\
&\quad + \left.\left(-\log P_{\text{ref}}(y_l|x,c) + \beta^{-1}\Delta\right) \cdot \mathbb{I}(y=y_l)\right] \\
&= \mathbb{E}\left[-\log P_{\text{ref}}(y_w|x,c) \cdot \mathbb{I}(y=y_w)\right] \\
& - \log P_{\text{ref}}(y_l|x,c) \cdot \mathbb{I}(y=y_l) \\
&\quad - \beta^{-1}\Delta \cdot \left(\mathbb{I}(y=y_w) - \mathbb{I}(y=y_l)\right) \\
&= \mathbb{E}\left[-\log P_{\text{ref}}(y_w|x,c) \cdot \mathbb{I}(y=y_w)\right] \\
&\quad + \mathbb{E}\left[-\log P_{\text{ref}}(y_l|x,c) \cdot \mathbb{I}(y=y_l)\right] \\
&\quad + \mathbb{E}\left[-\beta^{-1}\Delta \cdot \mathbb{I}(y=y_w) + \beta^{-1}\Delta \cdot \mathbb{I}(y=y_l)\right] \\
&= \mathcal{E}(\theta_{\text{SFT}}) + \beta^{-1} \mathbb{E}\left[\Delta \cdot \left(-\mathbb{I}(y=y_w) + \mathbb{I}(y=y_l)\right)\right]
\end{align*}

\noindent Given the uniform preference assumption $\mathbb{E}[\mathbb{I}(y=y_w)] = \mathbb{E}[\mathbb{I}(y=y_l)] = 0.5$, we have:
\begin{equation}
\nonumber
-\mathbb{I}(y=y_w) + \mathbb{I}(y=y_l) = \begin{cases}
-1 & \text{if } y=y_w \\
1 & \text{if } y=y_l
\end{cases} = -\frac{\mathbb{E}[\Delta]}{\beta}
\end{equation}

\noindent Thus, the final expression simplifies to:
\begin{equation}
\nonumber
\mathcal{E}(\theta_{\text{DPO}}) \leq \mathcal{E}(\theta_{\text{SFT}}) - \frac{\mathbb{E}[\Delta]}{\beta}
\end{equation}

\noindent As shown in the above equation, after preference alignment optimization through DPO, the preference strength $\mathbb{E}[\Delta]$ increases. Consequently, the upper bound of $\mathcal{E}(\theta_{\text{DPO}})$, given by $\mathcal{E}(\theta_{\text{SFT}}) - \frac{\mathbb{E}[\Delta]}{\beta}$, decreases. This reduction further lowers the generalization error, with an improvement magnitude of $\frac{\mathbb{E}[\Delta]}{\beta}$.

\noindent In summary, the joint training framework of GraphRAG, SFT, and DPO yields a total error upper bound expressed as:
\begin{equation}
\nonumber
\mathcal{E}(\theta_{\text{Final}}) \leq \mathcal{E}(\theta_{\text{SFT}}) - \frac{\gamma}{\beta} - \frac{\mathbb{E}[\Delta]}{\beta}
\end{equation}

\noindent Since the integration of GraphRAG enhances the information gain $\gamma$, which directly reduces the upper bound on the generalization error $\mathcal{E}(\theta_{\text{SFT+GraphRAG}})$. After preference alignment optimization through DPO, the preference strength $\mathbb{E}[\Delta]$ increases. Consequently, the upper bound of $\mathcal{E}(\theta_{\text{DPO}})$, given by $\mathcal{E}(\theta_{\text{SFT}}) - \frac{\mathbb{E}[\Delta]}{\beta}$, decreases.

\noindent  \subsubsection{Proposition 3}\textbf{ Reduced Hallucination in GraphRAG+SFT} \ \ 
When the retrieved context $c= \pi_\theta(x, G_x^*)$ from GraphRAG covers the relevant facts of $x$ with probability $P(F_x \subseteq c(x, G)) \geq 1 - \varepsilon$, the hallucination probability $P_{\text{hall}}(y|x, c(x, G))$ is bounded by $\varepsilon + \delta$, i.e.,
\begin{equation}
P_{\text{hall}}(y|x, c(x, G)) \leq \varepsilon + \delta,
\end{equation}
where $\varepsilon$ represents the training error, reflecting retrieval quality (lower $\varepsilon$ indicates better retrieval accuracy), and $\delta$ denotes the model's probability of relying on the retrieved content (smaller $\delta$ indicates stronger trust in the retrieval results).

\noindent \textbf{Proof:}
\begin{align*}
P_{\text{hall}} (y \mid x, c(x,G)) &= P_{\text{hall}} (y \mid x, c) \\
&= \sum_{y \notin F} P_\theta (y \mid x, c) \\
&= \sum_{y \notin F_x} P_\theta (y \mid x, c) + \sum_{y \notin F \setminus F_x} P_\theta (y \mid x, c) \\
&\leq \delta \cdot I(F_x \subseteq c) + \varepsilon \quad (\because P_\theta (F_x \not\subseteq c) \leq \varepsilon) \\
&\leq \delta + \varepsilon \quad (\because \delta \cdot I(F_x \subseteq c) \leq \delta)
\end{align*}

\noindent After minimizing $\mathcal{L}_{\text{GraphRAG+SFT}}$, the values of $\delta$ and $\varepsilon$ decrease. Consequently, combining GraphRAG with SFT reduces the upper bound $\delta + \varepsilon$ on model-generated hallucinations, thereby lowering the probability of hallucination occurrences.

\noindent  \subsubsection{Proposition 4}\textbf{ Reduced Hallucination in GraphRAG with SFT and DPO} \ \  When combining GraphRAG with SFT and further optimizing through DPO, the model's probability of generating hallucinated outputs $y_l$ is reduced. This probability is upper-bounded by $P_{\text{ref}} (y_l |x,c) \cdot e^{-\beta^{-1}\Delta}$, where $\Delta = \log \frac{P_{\text{ref}} (y_w |x,c)}{P_{\text{ref}} (y_l |x,c)}$.

\noindent \textbf{Proof:} From the proof of Proposition 2, we have known the optimal solution of DPO:
\begin{equation}
\nonumber
P_\theta (y|x,c) = \frac{1}{Z(x,c)} P_{\text{ref}} (y|x,c) e^{\beta^{-1} r(x,y)},
\end{equation}

\noindent  where $Z(x,c)$ represents the normalization factor. Substituting $Z(x,c)$ in the implicit reward function $r(x,y) = \beta \log \frac{P_\theta (y|x,c)}{P_{\text{ref}} (y|x,c)}$, we have:
\begin{equation}
\nonumber
P_\theta (y|x,c) = \frac{1}{Z(x,c)} P_{\text{ref}} (y|x,c) \cdot \frac{P_\theta (y|x,c)}{P_{\text{ref}} (y|x,c)} = \frac{P_\theta (y|x,c)}{Z(x,c)}.
\end{equation}
\noindent  This indicates that the optimal solution satisfies:
\begin{equation}
\nonumber
Z(x,c) = 1 \implies P_\theta (y|x,c) = P_{\text{ref}} (y|x,c) e^{\beta^{-1} r(x,y)}.
\end{equation}

\noindent  For hallucinated responses $y_l$, DPO suppresses their implicit reward $r(x,y_l)$. This effect can be derived from the DPO objective function $\mathcal{L}_{\text{DPO}}$ (i.e., Eq~\eqref{dpo}).  Minimizing $\mathcal{L}_{\text{DPO}} (\pi_\theta, \pi_{\text{ref}})$ requires maximizing the ratio $\frac{\pi_\theta (y_w|x)}{\pi_{\text{ref}} (y_w|x)}$ and minimizing the ratio $\log \frac{\pi_\theta (y_l|x)}{\pi_{\text{ref}} (y_l|x)}$. Since $\pi_{\text{ref}} (y_w|x)$ and $\pi_{\text{ref}} (y_l|x)$ remain fixed during optimization, this corresponds to increasing the implicit reward $r(x,y_w)$ for preferred outputs and decreasing the implicit reward $r(x,y_l)$ for hallucinated outputs. Assume the implicit rewards satisfy:
\begin{equation}
\nonumber
r(x,y_w) \geq r(x,y_l) + \Delta,
\end{equation}
\noindent  where $\Delta > 0$ represents the preference strength that increases during DPO training. Through gradient analysis of the DPO objective $\mathcal{L}_{\text{DPO}} (\pi_\theta, \pi_{\text{ref}})$ and setting the gradient to zero yields the optimal policy ratio, we obtain:
\begin{equation}
\nonumber
\frac{P_\theta (y_w|x,c)}{P_\theta (y_l|x,c)} = \frac{P_{\text{ref}} (y_w|x,c)}{P_{\text{ref}} (y_l|x,c)} \exp\left(\beta^{-1} (r(x,y_w) - r(x,y_l))\right).
\end{equation}

\noindent  Taking logarithms and applying the reward difference bound, we obtain:
\begin{equation}
\nonumber
\log \frac{P_\theta (y_w|x,c)}{P_\theta (y_l|x,c)} = \log \frac{P_{\text{ref}} (y_w|x,c)}{P_{\text{ref}} (y_l|x,c)} + \beta^{-1} (r(x,y_w) - r(x,y_l)).
\end{equation}

\noindent  Take the exponent and substitute $r(x,y_w) - r(x,y_l) \geq \Delta$, we obtain:
\begin{equation}
\nonumber
\frac{P_\theta (y_w|x,c)}{P_\theta (y_l|x,c)} \geq \frac{P_{\text{ref}} (y_w|x,c)}{P_{\text{ref}} (y_l|x,c)} \cdot e^{\beta^{-1} \Delta}.
\end{equation}

\noindent  Thus,
\begin{equation}
\nonumber
P_\theta (y_l|x,c) \leq P_{\text{ref}} (y_l|x,c) \cdot \frac{e^{\beta^{-1} \Delta} P_{\text{ref}} (y_w|x,c)}{P_\theta (y_w|x,c)}.
\end{equation}

\noindent  Given that the optimized policy maintains approximate fidelity to the reference model for preferred outputs (i.e., $P_\theta (y_w|x,c) \approx P_{\text{ref}} (y_w|x,c)$), where $P_{\text{ref}} (y_w|x,c)$ denotes the positive example probability initialized by SFT, we derive the approximate bound:
\begin{equation}
P_\theta (y_l|x,c) \leq P_{\text{ref}} (y_l|x,c) \cdot e^{-\beta^{-1} \Delta}.
\end{equation}

\noindent Since the function $e^{-\beta^{-1} \Delta}$ is strictly decreasing in $\Delta$, the upper bound on hallucination probability $P_{\text{ref}} (y_l|x,c) \cdot e^{\beta^{-1} \Delta}$ decays exponentially as $\Delta$ grows. The decay rate is governed by $-\beta^{-1} \Delta$, implying stronger preference signals (larger $\Delta$) yield exponentially tighter bounds and the inverse temperature parameter $\beta$ modulates the suppression intensity.

\section{Experiments}

We perform comprehensive experiments to answer the following research questions:

\noindent \textbf{RQ1}: Does~\sRC outperform the baseline methods in terms of overall performance?

\noindent \textbf{RQ2}: Can GraphRAG and fine-tuning LLMs in~\sRC improve performance?

\noindent \textbf{RQ3}:  What is the impact of choosing different top-$k$ values during the local and global information retrieval processes? 

\noindent \textbf{RQ4}:  During the fine-tuning stage, can incorporating generative pre-training (GPT) enhance performance?

\noindent \textbf{RQ5}: Does the selection of different base models affect the performance of~\RC?

\noindent \textbf{RQ6}:  What is the total time required by the model during the inference stage, encompassing both the GraphRAG retrieval time and the answer generation time?

\noindent  \textbf{RQ7}: How does~\sRC perform on clinical data in comparison to other baseline methods?

\subsection{ Experimental Settings}
We select the dataset presented in  Section~\ref{subsec:collect_dataset} to construct the knowledge graph and TCM formula instructions, and randomly choose 60,000 data pieces to perform supervised fine-tuning. Given a symptom $x$, we require the supervised fine-tuned LLM $ \pi_{\text{ref}} $ to generate 3,000 pairs of answers and score them as $\{y_w, y_l\}$, which are then used to execute DPO. Additionally, we randomly select 5,000 data pieces for model testing. We use Llama3-Chinese-Chat~\cite{Dubey-Arxiv2024-23} as the base model, use two NVIDIA RTX 4090 GPUs (24GB each). Memory usage was 32GB for GraphRAG alone, 36GB with GraphRAG+SFT, and 40GB for the full GraphRAG+SFT+DPO pipeline. We adopt LoRA as the fine-tuning strategy. We set the learning rate as 0.1, the  batch size as 1, the gradient accumulation steps as 8, and the maximum sequence length as 8,192 tokens. Additionally, we incorporate real-world clinical consultation data from Haodf.com~\footnote{https://challenge.xfyun.cn/topic/info?type=disease-claims-2022\&option=ssgy}, comprising 22,800 training and 7,600 test samples. Each record contains the patient’s age group (age), chief complaint (disease Name), consultation title (title), desired assistance (hope Help), descriptive text (condition Desc), and a visit direction label ($i \in$ [0, 19]). A secondary disease direction label ($j \in$ [0, 60]) is also included, with missing training labels marked as -1 (absent in the test set). Notably, we have obtained the ground-truth TCM formulas of the clinical dataset from TCM experts.
To evaluate zero-shot performance on clinical data, we directly prompt~\sRC to generate TCM formulas without additional training. Notably, to address conflicting knowledge, we develop specialized instruction datasets containing contradictory information scenarios, including Differences in Medical Theories, Conflicting Information Sources and Practical Problems. These datasets train the model to identify potential conflicts and generate appropriate warning messages, covering the following situations (See appendix for more details).

To evaluate the performance of the models, in additional to using traditional machine translation metrics (e.g., BLEU and ROUGE-1/2/L), we further carefully design six TCM-oriented quantitative evaluation metrics, i.e., \textit{Compatibility Compliance Rate} (CCR), \textit{{Correct Sovereign-minister-assistant-messenger Compatibility  Rate}} (CSCR), \textit{Counter Coarse Hallucination Rate} (CCHR), \textit{FactScore} (FS)\footnote{Alternatively termed as Counter Fine-grained Hallucination Rate. }, \textit{Structural/Content Clarity Rate} (SCR)  and \textit{Logical Rate} (LR) to enhance the transparency and robustness of the assessment. Table~\ref{tb:quantity} illustrates these six indicators. Each indicator is scored on a 0–1 scale, with higher values indicating better performance. Notably, due to space constraints, we employ the notation ROUGE-S to denote the averaged ROUGE-1/2/L scores.  Besides, we evaluate computational efficiency by reporting training FLOPs in Table~\ref{tb:overall}-~\ref{tb:clinical}, which quantify each model's computational complexity.

\begin{table*}
	\centering
	\caption{ Six TCM-oriented quantitative indicators.}
	\label{tb:quantity}
	\small
	\begin{tabular}{p{3.6cm}|p{6.2cm}|p{6.2cm}p{0.00001cm}}
		\toprule
		\centering Name &\centering Definition&  \centering Formula&   \\ 
		\midrule 
		Compatibility Compliance Rate (CCR) & Check if the generated herbal formula breaks any TCM compatibility rules, like the `18 Incompatible' or `19 Antagonistic' herb pairs. &CCR = (1-   $\frac{  {\rm Number  \ of \ non-compliant \  herb \ pairs } }{{\rm Total \ number \ of\  herb \ pairs }}$) $\times 100\%$ \\ \midrule
		Correct Sovereign-minister-assistant-messenger Compatibility  Rate (CSCR)
		& Evaluate whether the prescription's herb roles (sovereign, minister, assistant, messenger) follow  TCM principles and clinical evidence.		& CSCR = $w_sr_s+w_{mi}r_{mi}+w_ar_a+w_{me}r_{me}$, where $w_s$, $w_{mi}$, $w_a$, $w_{me}$ denote weights, $r_s$, $r_{mi}$, $r_a$, $r_{me}$ denote the compatibility accuracy rates for sovereign, minister, assistant, and envoy herbs, respectively. Without loss of generality, we set $w_s$, $w_{mi}$, $w_a$, $w_{me}$ as 0.25. &  \\ \midrule
		Counter Coarse Hallucination Rate (CCHR) & The likelihood that the model’s generated responses are factually accurate. & CCHR = (1- $\frac{{\rm Number \ of \ hallucinated \ responses}}{{\rm Number \ of  \ total \ response}}$$\times 100\%$) \\ \midrule
		FactScore (FS)& A fine-grained evaluation metric to quantify the factual accuracy of generative model outputs at the atomic level~\cite{Min2023Fact}.& FS = $\frac{{\rm Number \ of \ supported \ facts} }{{\rm Number \ of \ total \ asserted \ facts}}$$\times 100\%$\\ \midrule
	    Structural/Content Clarity Rate (SCR)&  SCR consists of Content Clarity Rate (CR) and Content Professional Rate (CPR). $a$ denotes the exact count of six key output components: (1) recommended prescriptions, (2) TCM ingredients, (3) applicable population symptoms, (4) pulse and tongue coating characteristics, (5) prescription contraindications, and (6) processing methods.
	     &SCR= 0.5 $\times $CR  + 0.5$\times$ CPR; 
	     CR = $\frac{a}{6} $ $\times 100\%$;  
	     CPR=$\frac{{\rm Number \ of  \ sentences  \ meeting  \ professional \ criteria}}{ {\rm Number \ of \ all \ sentences}}$ $\times 100\%$ 
	     & \\ \midrule
		Logical Rate (LR)& Evaluate whether each contextual component of the answer demonstrates a coherent logical relationship. &LR=$\frac{{\rm Number \ of \ coherent  \ contextual  \ sentences}}{{\rm Number \ of \ contextual \ sentences}}$ $\times 100\%$    &    \\
		
		\bottomrule
	\end{tabular}
\end{table*}

\hide{Furthermore, to assess the quality of TCM formulas and their explanations, we establish an accuracy metric called \textit{Precision}, which evaluates whether the the composition of the formula's sovereign, minister, assistant, courier; efficacy; contraindications; tongue and pulse diagnosis predicted by each model are accurate.
Additionally, we introduce a comprehensive evaluation metric called \textit{Composite Score} (CS), which includes ten criteria: comprehensiveness, logic, interpretability, scientific validity, practicality, consistency, relevance, language expression, risk assessment, and safety. Each criterion is scored out of 10 points, with a total possible score of 100 points.
}

\subsection{ Baselines}
Since traditional TCM formula models~\cite{JingCNKI2024-1,LiuTCM2017-2,NimaTCM2021-3} only mine the associations between any two herbs in a formula, and Zhou et al.~\cite{XueTCMLLM-37} have demonstrated that fine-tuning LLMs~\cite{XueTCMLLM-37} can outperform the state-of-the-art deep learning model TCMPR~\cite{DongBio35}. Therefore,  our comparative analysis focuses exclusively on LLM-based TCM formula models. As detailed in Section~\ref{subsec:problem_definition}, we 
evaluate six baseline approaches representing different methodological paradigms: 1) Plug-and-play, 2) FT, 3) RAG, 4) GraphRAG, 5) RAG+FT and 6) GraphRAG+SFT architectures.


\noindent 1) LLaMA~\cite{Dubey-Arxiv2024-23}, Kimi~\cite{Kimi-27}, DeepSeek~\cite{Deepseek-28}, Qwen~\cite{Qianwen48}: Plug-and-play approaches using LLaMA3.2-7B, Kimi-7B, DeepSeek-7B, and Qwen2.5-7B to directly output TCM formulas and explanations.

\noindent 2) TCMLLM~\cite{XueTCMLLM-37}: A FT approach, which constructs a TCM formula instruction dataset to fine-tune LLaMA and output TCM formulas and explanations. The constructed dataset only contains the composition of the formula's sovereign, minister, assistant, courier based on symptoms.

\noindent 3) RAG~\cite{XiongACL2024}: A RAG approach using Llama3.2-7B as the base model. For efficient similarity search, we utilize FAISS (Facebook AI Similarity Search\footnote{https://github.com/facebookresearch/faiss}), an embedding matching library implemented in LlamaIndex, to retrieve the top-k results. Notably, all following methods involving RAG use the above experimental settings.  

\noindent 4) GraphRAG~\cite{Edge-Arxiv-26}: A GraphRAG approach using Llama3.2-7B as the base model. 
Perform local retrieval in documents based on symptoms and aggregate local retrieval information to obtain global retrieval content, then output TCM formulas and explanations based on symptoms and the global retrieval content.  In practice, we also use FAISS to retrieval top-$k$ relevant chunks.

\noindent 5) RAG+Fine-tuning:  A RAG+FT approach using Llama3.2-7B as the base model and fine-tune it, where the fine-tuning dataset is constructed by RAG. The process includes constructing a TCM formula instruction dataset for fine-tuning, involving both SFT and DPO processes.  

\noindent 6) RAG+SFT~\cite{Giuffre2025DLD}: A RAG+FT approach using Llama3.2-7B as the base model that only performs SFT during the fine-tuning step, where the fine-tuning dataset is constructed by RAG. 

\noindent 7) \RC: Our proposed model implements a GraphRAG+FT approach, utilizing Llama3.2-7B as the base model. The framework integrates GraphRAG with SFT and DPO techniques.

\noindent 8) GraphRAG+SFT: A GraphRAG+FT approach A method that only performs SFT during the fine-tuning step of the~\sRC model.

For all retrieval-augmented methods (RAG, GraphRAG, RAG+Fine-tuning, RAG+SFT, GraphRAG+SFT, and~\sRC), we consistently set top-$k$ as 3.

\subsection{ Experimental Results}
\subsubsection{Overall Results}
To explore whether~\sRC outperforms the baseline methods in terms of overall performance (RQ1), we set top-$k$ as 3  and evaluated each model on the test set. The results are shown in Table~\ref{tb:overall}. The results indicate that: 1) \sRC achieves the highest performance, followed by GraphRAG+SFT. This indicates that integrating GraphRAG with SFT reduces generalization error (cf. BLEU, ROUGE-S, CCR, etc) and hallucination rates (cf. FS and SCR) in TCM formula analysis. Furthermore, incorporating DPO further decreases these errors.
2) Combining SFT with retrieval-based methods (e.g., GraphRAG, RAG) leads to a significant performance improvement, particularly in CCHR and FS metrics. Notably,~\sRC  achieves the highest CCHR and FS scores, indicating that integrating GraphRAG with SFT effectively reduces both generalization error and hallucination rates.
3) LLaMA performs the worst, which implies directly using vanilla LLM is not sufficient to solve the domain-specific downstream tasks. 
4) Compared to vanilla LLM models (i.e., LLaMA, Kimi, Qwen),  using RAG/GraphRAG alone or fine-tuning LLaMA alone can achieve initial performance improvements, which demonstrates the effectiveness of using RAG/GraphRAG or  fine-tuning techniques.
5) RAG performs worse than GraphRAG, which demonstrates that GraphRAG can retrieve fine-grained external information.
6) Although~\sRC incurs higher training FLOPs than the baselines—implying longer training time—its inference FLOPs remain only marginally above those of direct API use (e.g., $\sim$1.68$\cdot10^5$TFLOPs vs $\sim$1.21$\cdot10^5$TFLOPs), because the GraphRAG retrieval stage is lightweight.

\begin{table*}
	\centering
	\caption{Overall performances, top-$k$=3. Term ``Avg" refers to the mean score across all metrics.}
	\label{tb:overall}
	\small
	\begin{tabular}{|p{2.4cm}|p{1.0cm}|p{1.4cm}|p{1.0cm}|p{1.0cm}|p{1.0cm}|p{0.8cm}|p{1.0cm}|p{0.8cm}|p{0.8cm}|p{1.6cm}|p{0.001cm}}
		\toprule
		\centering Model &\centering BLEU&\centering ROUGE-S&\centering CCR&  \centering  CSCR &\centering CCHR&\centering FS&\centering SCR &\centering LR &\centering Avg &\ \ \ FLOPs\\ 
		\midrule 
		\centering LLaMA & \centering 0.0013& \centering	0.0398& \centering 0.8720		&\centering 0.8379& \centering	0.8721&\centering 0.9236 	&\centering 0.7569&\centering 0.8634 & \centering 0.6459&\textbf{$\sim$1.21$\cdot\textbf{10}^\textbf{5}$T}  \\ 
		\centering Qwen  & \centering 0.0204 &\centering 	0.0213	 & \centering 0.8693  &\centering 0.8412		& \centering 0.8640 	&  \centering 0.9063 &\centering 0.7524&\centering 0.8637 & \centering 0.6416&\textbf{$\sim$1.21$\cdot\textbf{10}^\textbf{5}$T} \\ 
		\centering Kimi &\centering 0.0221  &\centering	0.3962&	\centering 0.8940 &\centering 0.8639&\centering 0.8908 &\centering	0.9315& \centering 0.8540& \centering 0.9045 &\centering 0.7196&\textbf{$\sim$1.21$\cdot\textbf{10}^\textbf{5}$T} 	 \\ 
		\centering DeepSeek &  \centering 0.0233&\centering	0.3004&	\centering 0.8806 &\centering 0.8574 &	\centering 	0.8890& \centering 0.9318& \centering 0.7806 &\centering 0.8732 &\centering 0.5536 &\textbf{$\sim$1.21$\cdot\textbf{10}^\textbf{5}$T} \\  \midrule
		\centering TCMLLM  &   \centering 0.0106 &  \centering 0.0064	&   \centering 0.8724& 	 \centering 0.8370	& \centering 0.8720 	&  \centering 0.9240 & \centering 0.7547 & \centering 0.8624 &\centering 0.6425 &$\sim$2.30$\cdot10^{6}$T \\  \midrule
		\centering RAG  &  \centering  0.0778  & \centering 0.2838	&	 \centering  0.8659& \centering 	0.8641&   \centering 0.8890 & \centering 0.9330	& \centering 0.7805 &\centering 0.8732 & \centering 0.6959 & $\sim$1.21$\cdot10^5$T \\ 
		\centering GraphRAG &  \centering  0.0876 & \centering  0.3152 & \centering  0.8827&	 \centering  0.8973	& \centering 0.9151 	& \centering 0.9506 & \centering 0.8506 &\centering 0.9031 &\centering 0.7252 &$\sim$1.68$\cdot10^5$T\\  \midrule
		\centering RAG+Fine-tuning &  \centering 0.1216	& \centering 0.3403 &	\centering 0.8718 &	\centering  0.8721& \centering 0.9015& \centering 0.9444&\centering 0.8485 &\centering 0.9022 & \centering 0.7253 &$\sim$7.59$\cdot10^{6}$T  \\ 
		\centering RAG+SFT  &\centering 0.3641&	\centering 0.5377&\centering 0.8804	&	\centering 0.8959&\centering 0.9113 	&	\centering 0.9436 & \centering 0.9080 & \centering 0.9301 & \centering 0.7964 &$\sim$3.51$\cdot10^{6}$T \\ \midrule
		\centering GraphRAG+SFT& \centering 0.3921	&\centering 0.6388	& \centering 0.8847	& \centering  0.8950&	\centering 0.9126 &\centering 0.9505 &\centering 0.9174 &\centering 0.9342 & \centering 0.8157 &$\sim$3.98$\cdot10^{6}$T \\ \midrule
		\centering  \textbf{\RC}& \centering \textbf{0.4213}	& \centering  \textbf{0.6667}	&\centering  \textbf{0.9383}	&\centering  \textbf{0.9062}	&\centering \textbf{0.9303}&	\centering \textbf{0.9741}  &\centering  \textbf{0.9248} & \centering  \textbf{0.9374} &\textbf{0.8373} &$\sim$8.06$\cdot10^{6}$T  \\
		\bottomrule
	\end{tabular}
\end{table*}

Additionally, to investigate whether GraphRAG and fine-tuning LLMs in~\sRC can improve performance (RQ2), we have the following findings based on Table~\ref{tb:overall}:
1) GraphRAG and fine-tuned TCMLLM both surpass vanilla LLaMA in TCM formula analysis, with GraphRAG showing superior gains. While both methods improve accuracy and interpretability, TCMLLM's limitation to sovereign/minister/assistant/courier components restricts its fine-grained analysis capability.
2) Although LLaMA underperforms Qwen, Kimi, and DeepSeek, all enhanced approaches (RAG/SFT variants and~\RC) outperform these models. This implies that GraphRAG combined with fine-tuning effectively compensates for base model deficiencies.

\begin{figure}[t]
	\centering
	\includegraphics[width= 0.43 \textwidth]{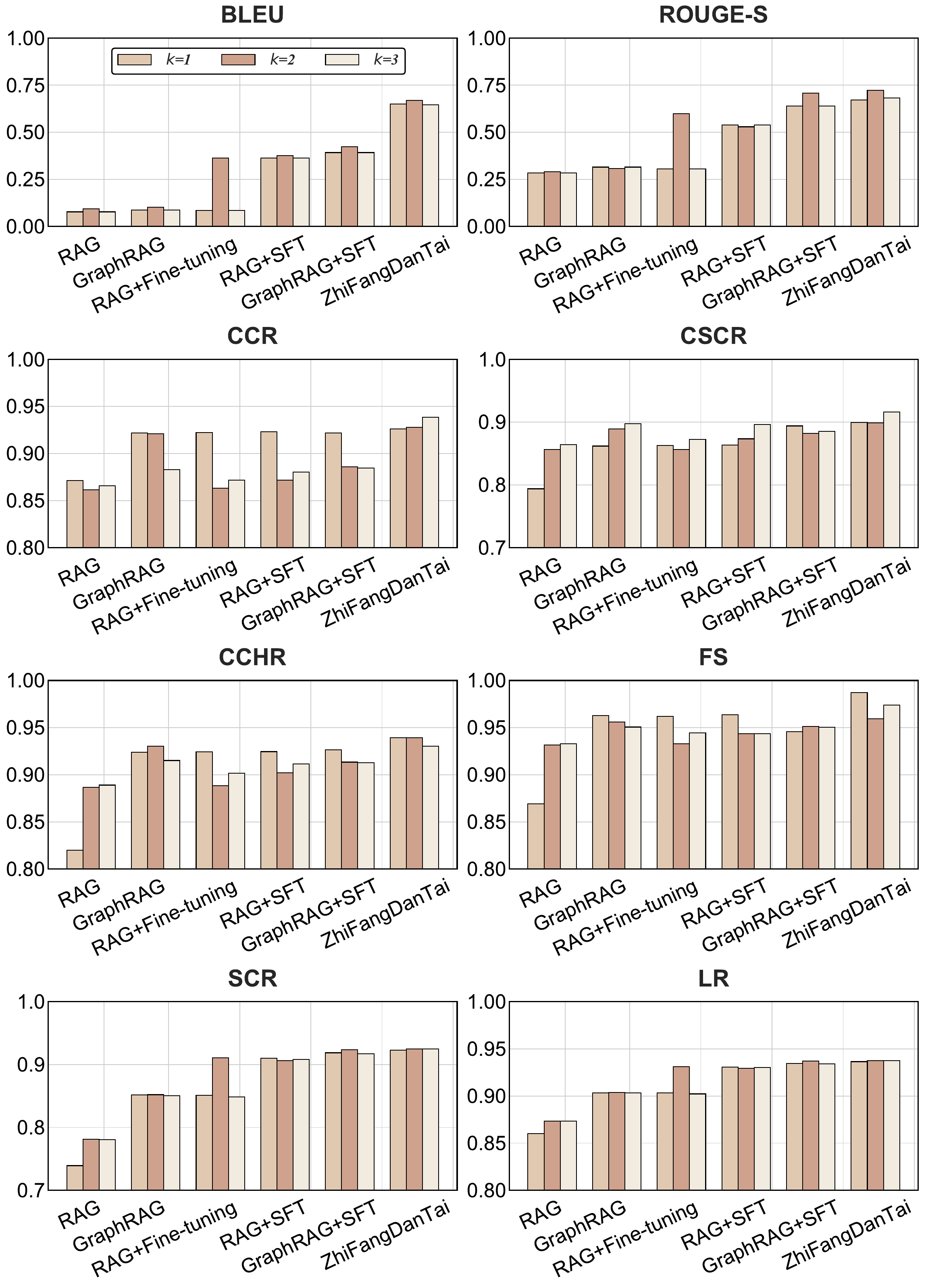}
	\caption{\label{fig:topk}   Performance of each method under different  top-$k$.}
\end{figure}
\begin{table*}
	\centering
	\caption{Comparison of GPT, SFT and DPO, top-$k$=3. ``Avg" refers to the mean score across all metrics.}
	\label{tb:generative_pre_training}
	\small
	\begin{tabular}{|p{2.8cm}|p{1.0cm}|p{1.4cm}|p{1.0cm}|p{1.0cm}|p{1.0cm}|p{0.8cm}|p{1.0cm}|p{0.8cm}|p{0.8cm}|p{1.5cm}|p{0.001cm}}
		\toprule
		\centering Model &\centering BLEU&\centering ROUGE-S&\centering CCR&  \centering CSCR &\centering CCHR&\centering FS&\centering SCR &\centering LR & \centering Avg &\ \ \ FLOPs\\  
		\midrule 	\centering GraphRAG+GPT &	\centering 0.0330& 	\centering 	0.2670& 	\centering 0.8632 	& 	\centering 	0.8232& 	\centering 0.8605 	& 	\centering 0.9237 & \centering 0.7554&\centering 0.8650 &\centering 0.6749& $\sim$1.01$\cdot10^{7}$T\\\midrule
		\centering GraphRAG+GPT+SFT &	\centering  0.3858& 	\centering 	0.5693& 	\centering 0.8947 	& 	\centering 	0.8796& 	\centering 0.9137 	& 	\centering 0.9535 & \centering 0.9093&\centering 0.9282 &\centering 0.8043&$\sim$1.24$\cdot10^{7}$T \\\midrule
		\centering GraphRAG+GPT+SFT +DPO & \centering0.3928	& \centering 0.5969& \centering 0.8950	& \centering	0.8831& \centering	0.9225& \centering	0.9536&\centering 0.9133 & \centering 0.9298& \centering 0.8109&$\sim$1.64$\cdot10^{7}$T\\  \midrule
	\centering  \textbf{\RC} & \centering \textbf{0.4412}& \centering 	\textbf{0.7181}	& \centering \textbf{0.9357}	& \centering \textbf{0.9724}& \centering \textbf{0.9523}& \centering \textbf{0.9671} & \centering \textbf{0.8978} & \centering \textbf{0.9253}& \centering \textbf{0.8512} &\textbf{$\sim$8.06$\cdot\textbf{10}^{\textbf{6}}$T}\\
		\bottomrule
	\end{tabular}
\end{table*}

\begin{table*}
	\centering
	\caption{Overall Performance when using Qwen as base model, top-$k$=3. ``Avg" refers to the mean score across all metrics.}
	\label{tb:qwen}
	\small
	\begin{tabular}{|p{2.4cm}|p{1.0cm}|p{1.4cm}|p{1.0cm}|p{1.0cm}|p{1.0cm}|p{0.8cm}|p{1.0cm}|p{0.8cm}|p{0.8cm}|p{1.6cm}|p{0.001cm}}
		\toprule
\centering Model &\centering BLEU&\centering ROUGE-S&\centering CCR&  \centering CSCR &\centering CCHR&\centering FS&\centering SCR &\centering LR &\centering  Avg &\ \ \ FLOPs \\ 
		\midrule \centering Qwen
& \centering 0.0204 &\centering 0.3083	 & \centering 0.8793  &\centering 0.8412		& \centering 0.8740 	&  \centering 0.9263 &\centering 0.7724&\centering 0.8732 & \centering 0.6869&  \textbf{$\sim$1.21$\cdot\textbf{10}^\textbf{5}$T}\\ \midrule
		\centering TCMLLM &\centering  0.0129&\centering 0.1098&  \centering 0.8774& 	 \centering 0.8392	& \centering 0.8753 	&  \centering 0.9282 & \centering 0.7577 & \centering 0.8629 &\centering 0.6579 &$\sim$2.30$\cdot10^{6}$T \\ \midrule
		\centering RAG &  \centering  0.1277&  \centering 0.4326&  \centering 0.8632	&  \centering  0.8921	&  \centering 0.9001&  \centering 0.9419&\centering 0.7989 &\centering 0.8813 & 0.7297 &$\sim$1.21$\cdot10^{6}$T \\ 
		\centering GraphRAG&  \centering 0.1079	&  \centering 0.4601&  \centering 0.8930		&  \centering 0.9013&  \centering 0.9402	&  \centering  0.9621&\centering 0.8006&\centering 0.8821 & 0.7434 &$\sim$1.68$\cdot10^{6}$T \\  \midrule
		\centering RAG+Fine-tuning	&  \centering 0.2038	&  \centering 	0.4953	&  \centering 0.8801		&  \centering 	0.8624	&  \centering 0.9193 	&  \centering 0.9536 & \centering 0.8882 &\centering 0.9205 &0.7653 &$\sim$7.59$\cdot10^{6}$T   \\ 
		\centering RAG+SFT&\centering  0.1760&\centering 0.5837	&\centering 0.8696	&\centering 0.8698	&\centering 0.9212&\centering 0.9589 	&\centering 0.8904 & \centering 0.9214 & 0.7739&$\sim$3.51$\cdot10^{6}$T\\ \midrule
		\centering GraphRAG+SFT & \centering0.4372 & \centering 0.7059& \centering	0.9124& \centering 0.8816	& \centering 0.9502& \centering 0.9623 &\centering 0.8900 &\centering0.9221 & 0.8327&$\sim$3.98$\cdot10^{6}$T \\ \midrule
		\centering  \textbf{\RC} & \centering \textbf{0.4412}& \centering 	\textbf{0.7181}	& \centering \textbf{0.9357}	& \centering \textbf{0.9724}& \centering \textbf{0.9523}& \centering \textbf{0.9671} & \centering \textbf{0.8978} & \centering \textbf{0.9253}& \centering \textbf{0.8512} &$\sim$8.06$\cdot10^{6}$T\\
		\bottomrule
	\end{tabular}
\end{table*}
\subsubsection{ Detailed analysis }
\label{subsec:topk}
To explore what is the impact  of choosing different  top-$k$  during the local and global information retrieval processes  (RQ3), we evaluate the models with different top-$k$ values, i.e., top-$k$= 1, 2, and 3, and report the results in Fig.~\ref{fig:topk}. The results show that: 1) Compared to all other models,~\sRC consistently achieves the best performance across different values of top-$k$. It shows only a marginal improvement in TCM-oriented metrics but a significant improvement in machine translation metrics.
This indicates that GraphRAG has highly accurate retrieval capabilities, making a top-$k$ value of 1 sufficient. As top-$k$ increases, the model retrieves a larger amount of reference content, resulting in more comprehensive answer coverage. Consequently, both BLEU and ROUGE-S scores improve significantly with higher top-$k$ values. 
2) For each model, using top-$k$=1 may occasionally degrade performance, as the limited number of retrieved references can impair the LLM's ability to generate accurate responses.  Since top-$k$=2 or 3 demonstrates optimal performance in most cases, we select top-$k$=2 for computational efficiency.

\begin{table*}
	\centering
	\caption{Clinical data performances, top-$k$=2. Term ``Avg" refers to the mean score across all metrics.}
	\label{tb:clinical}
	\small
	\begin{tabular}{|p{2.4cm}|p{1.0cm}|p{1.4cm}|p{1.0cm}|p{1.0cm}|p{1.0cm}|p{0.8cm}|p{1.0cm}|p{0.8cm}|p{0.8cm}|p{1.5cm}|p{0.001cm}}
		\toprule
		\centering Model &\centering BLEU&\centering ROUGE-S&\centering CCR&  \centering CSCR &\centering CCHR&\centering FS&\centering SCR &\centering LR &\centering Avg &\ \ \ FLOPs\\ 
		\midrule 
		\centering LLaMA & \centering 0.0015& \centering	0.0486& \centering 0.6647		&\centering 0.6048& \centering	0.8135&\centering 0.7704 	&\centering 0.7010&\centering 0.8350 & \centering 0.5549&\textbf{$\sim$1.21$\cdot\textbf{10}^{\textbf{5}}$T} \\ 
		\centering Qwen  & \centering 0.0042 &\centering 	0.0348	 & \centering 0.6543  &\centering 0.5931		& \centering 0.8214 	&  \centering 0.7792 &\centering 0.6934&\centering 0.8201 & \centering 0.5501& \textbf{$\sim$1.21$\cdot\textbf{10}^{\textbf{5}}$T}\\ 
		\centering Kimi &\centering 0.0623  &\centering	0.3870&	\centering 0.8720 &\centering	0.8421&\centering 0.8834 &\centering	0.9032& \centering 0.8543& \centering 0.9051 &\centering 0.7136&	\textbf{$\sim$1.21$\cdot\textbf{10}^{\textbf{5}}$T}\\ 
		\centering DeepSeek &  \centering 0.0665&\centering	0.3950&	\centering 0.8802 &\centering 0.8462 &	\centering 0.8792& \centering 0.8992& \centering 0.8571 &\centering 0.9075 &\centering 0.7163 &\textbf{$\sim$1.21$\cdot\textbf{10}^{\textbf{5}}$T} \\  \midrule
		\centering TCMLLM  &   \centering 0.0019 &  \centering 0.4135	&   \centering 0.8913& 	 \centering 0.8512	& \centering 0.8821 	&  \centering 0.9123 & \centering 0.8645 & \centering 0.9095 &\centering 0.7158 &$\sim$1.53$\cdot10^{6}$T  \\  \midrule
		\centering RAG  &  \centering  0.0021  & \centering 0.2246	&	 \centering  0.7480& \centering 	0.6485&   \centering 0.7380 & \centering 0.8145	& \centering 0.7031 &\centering 0.8460 & \centering 0.5906 &$\sim$1.21$\cdot10^{6}$T \\ 
		\centering GraphRAG &  \centering 0.0028 & \centering  0.2255 & \centering  0.7635&	 \centering 0.6575	& \centering 0.7595 	& \centering 0.8365 & \centering 0.7187 &\centering 0.8470 &\centering 0.6014 &$\sim$1.32$\cdot10^{6}$T  \\  \midrule
		\centering RAG+Fine-tuning &  \centering 0.4123	& \centering 0.5024 &	\centering 0.9318 &	\centering 0.8925& \centering 0.9014& \centering 0.9046&\centering 0.8824 &\centering 0.9106 & \centering 0.7922 &$\sim$5.46$\cdot10^{6}$T  \\ 
		\centering RAG+SFT  &\centering 0.4330&	\centering 0.5188&\centering 0.9420	&	\centering 0.9005&\centering 0.9285 	&	\centering 0.9324 & \centering 0.9075 & \centering 0.9290 & \centering 0.8115 & $\sim$2.74$\cdot10^{6}$T\\ \midrule
		\centering GraphRAG+SFT& \centering 0.4371	&\centering 0.5304	& \centering 0.9405	& \centering 0.9025&	\centering 0.9290 &\centering 0.9348 &\centering 0.9100 &\centering 0.9335 & \centering 0.8147 &$\sim$2.85$\cdot10^{6}$T\\ \midrule
		\centering  \textbf{\RC}& \centering \textbf{0.4519}	& \centering  \textbf{0.5432}	&\centering  \textbf{0.9440}	&\centering  \textbf{0.9305}	&\centering  \textbf{0.9340}&	\centering \textbf{0.9371}  &\centering  \textbf{0.9103} & \centering  \textbf{0.9393} &\centering \textbf{0.8237} &$\sim$5.57$\cdot10^{6}$T \\
		\bottomrule
	\end{tabular}
\end{table*}
\begin{table}
	\centering
	\caption{Inference time comparison, top-$k$ for~\sRC is 3. ``Avg" refers to the mean score across all metrics.}
	\label{tb:inference}
	\small
	\begin{tabular}{|p{2.3cm}|p{0.9cm}|p{1.1cm}|p{1.22cm}|p{1.3cm}|p{0.001cm}}
		\toprule
		\centering Model &\centering Avg&  \centering Retrieval Time (s) &\centering Generation Time (s)& Generate Tokens\\ 
		\midrule 
		\centering DeepSeek & \centering 0.7163 & \centering	\textbf{$\sim$0.60s} & \centering	$\sim$11.8s &\ 1,124.2 \\
		\centering Kimi  & \centering 0.7136 &\centering 	\textbf{$\sim$0.60s}	 & \centering \textbf{$\sim$10.1s} & \ 1,128.7 \\ \midrule
		\centering \RC-P  & \centering 0.7392 &\centering $\sim$0.66s& \centering $\sim$~10.9s&\  1,002.8 \\ 
		\centering ~\sRC  & \centering \textbf{0.8237} &\centering $\sim$0.86s	 & \centering $\sim$~11.5s & \ 1,108.6 \\ 
		
		\bottomrule
	\end{tabular}
\end{table}

Besides, to explore whether incorporating generative pre-training (GPT) during the fine-tuning stage can improve the performance (RQ4), We compare four variant models: GraphRAG+GPT, GraphRAG+GPT+SFT, GraphRAG+GPT+SFT+DPO, and ZhiFangDanTai (GraphRAG+SFT+DPO). For performing GraphRAG+GPT, we use riginal collected TCM dataset $p_{data (x)}$, for performing GraphRAG+GPT+SFT and GraphRAG+GPT+SFT+DPO, we use original collected TCM dataset $p_{data (x)}$ and data-augmented data $p_{task (x)}$. For performing~\RC, we only use data-augmented data $p_{task (x)}$. The results are shown in Table~\ref{tb:generative_pre_training}, which indicate that incorporating GPT  did not enhance performance but even led to a decline. 
Here we provide theoretical analysis: 
	Since GPT is the only variable factor, we focus our analysis on comparing GraphRAG+GPT versus GraphRAG+GPT+SFT. Other comparisons follow similar way.
	1) From the perspective of data distribution deviation, the goal of GraphRAG+GPT is minimizing $D_{\text{KL}}(p_{\text{data}} \parallel p_\theta)$, while the goal of GraphRAG+GPT+SFT is minimizing $D_{\text{KL}}(p_{\text{task}} \parallel p_\theta)$. We denote these two variant models as $p_\theta^{(\text{GPT+SFT})}$ and $p_\theta^{\text{SFT}}$. Since $p_{\text{data}} \neq p_{\text{task}}$, we obtain $	D_{\text{KL}}(p_{\text{task}} \parallel p_\theta^{(\text{GPT+SFT})}) > D_{\text{KL}}(p_{\text{task}} \parallel p_\theta^{\text{SFT}})$, 
	i.e., compared with GraphRAG+SFT, GraphrAG+GPT+SFT leads to greater distributional deviation in the model's outputs on downstream tasks, resulting in performance degradation.
2) From the perspective of risk of overfitting, Goar et al.~\cite{Goar2024} have pointed out the model's generalization error is bounded by the Rademacher complexity $\mathfrak{R}_N(\mathcal{F})$, i.e., 
$
\epsilon(\theta_{\text{Generalization Error}}) \leq \mathfrak{R}_N(\mathcal{F}) + \mathcal{O}\sqrt{\frac{\log(1/\delta)}{N}}$,
where $\mathcal{F}$ is the hypothesis class and $N$ is the data volume. 
Since the dataset size in GPT training is significantly smaller than the parameter count of large language models (e.g., 7B parameters), and the Rademacher complexity $\mathfrak{R}_N(\mathcal{F})$ remains high, the upper bound on the model's generalization error consequently increases, leading to performance degradation. Based on these findings, we recommend that for the training of~\RC, GPT is unnecessary, as it not only increases training costs but also negatively impacts model performance.

Additionally, to explore whether the choice of the base model affects the performance of~\sRC  (RQ5), we conducted experiments using the Qwen2.5-7B (Qwen) model under the 
experimental setting with top-$k$=3 for retrieval items. The results are shown in Table~\ref{tb:qwen}, which indicate that~\sRC achieves superior performance across most scenarios regardless of whether Qwen or LLaMA is used as the base model. Moreover, combining fine-tuning techniques with RAG/GraphRAG technology outperforms using RAG/GraphRAG or fine-tuning alone, demonstrating the universality of the proposed method.

\subsubsection{ Model inference analysis }
To examine the total inference time encompassing GraphRAG retrieval and answer generation (RQ6), we compare~\sRC with DeepSeek and Kimi using all clinical and collected test samples, deploying them via WebUI (C.f. Section~\ref{subsec:model_inference}) while evaluating competitors through their official platforms. We further test a pruned GraphRAG variant (\RC-P) by reducing by communities into two. Results in Table~\ref{tb:inference} show~\sRC achieves superior performance with competitive retrieval times; while marginally slower in retrieval, this trade-off is justified by its accuracy advantage. Conversely,~\RC-P demonstrates reduced performance due to coarser retrieval granularity, with all systems evaluated across mean metric scores ("AVG"), timing measures, and output lengths.

\subsubsection{Clinical Data Performance}
To assess~\sRC's performance on cold-start clinical data relative to baseline methods (RQ7), we conducted a comparative evaluation by directly prompting both~\sRC and baseline models to generate TCM formulas from the given clinical records. The results, presented in Table~\ref{tb:clinical}, indicate that~\sRC achieves the highest performance and the lowest hallucination rate. This demonstrates~\RC's effectiveness in zero-shot learning scenarios, suggesting its potential for direct application to clinical datasets and utility as an AI assistant in medical settings.

\section{Conclusion}

In this paper, we propose~\RC, a  framework combining Graph-based Retrieval-Augmented Generation (GraphRAG) with LLM fine-tuning. ~\sRC uses GraphRAG to retrieve and synthesize structured TCM knowledge into concise summaries, while also constructing an enhanced instruction dataset to improve LLMs' ability to integrate retrieved information. 
	Furthermore, we provide novel theoretical proofs demonstrating the superiority of integrating GraphRAG with LLM fine-tuning. Experimental results on both collected and clinical datasets demonstrate that ~\sRC achieves significant improvements over state-of-the-art models in all carefully designed quantitative metrics.


\begin{IEEEbiography}[{\includegraphics[width=1in,height=1.25in,clip,keepaspectratio]{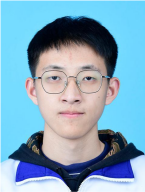}}]{Zixuan Zhang} is currently an undergraduate in Capital Normal University. His main research interests are large language model for medical diagnosis and multi-objective optimization. 
\end{IEEEbiography}

\begin{IEEEbiography}[{\includegraphics[width=1in,height=1.35in]{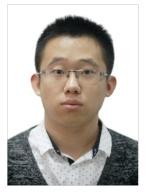}}]{Bowen Hao} reveived PhD degree in School of Information, Renmin University of China in 2022. He is currently an assistant professor in Capital Normal University. His research interests include data mining, large language model for recommendation, and
large language model for medical diagnosis. He has published 10+ papers in the top Data mining (TOIS, WSDM, AAAI) and Machines Learning (ECML-PKDD, IJIS, ApWeb-WAIN) venues.
\end{IEEEbiography}

\begin{IEEEbiography}[{\includegraphics[width=1in,height=1.35in,clip,keepaspectratio]{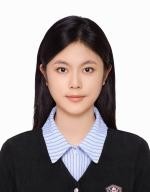}}]{Yingjie Li} is currently an undergraduate in Capital Normal University. Her main research interests are large language model for medical diagnosis and data mining. 
\end{IEEEbiography}

\begin{IEEEbiography} [{\includegraphics[width=1.05in,height=1.85in,clip,keepaspectratio]{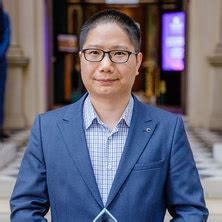}}]{Hongzhi Yin} (Senior Member, IEEE) reveived  PhD degree in computer science from Peking University, in 2014. He works as an ARC Future fellow and full professor with The University of Queensland, Australia. He has made notable contributions to recommendation systems, graph learning, and decentralized edge intelligence. He has published top 300+ papers ( KDD, SIGIR, WWW, ACL, WSDM, SIGMOD, VLDB, ICDE, NeurIPS, AAAI, IJCAI, ACM Multimedia, ECCV, IEEE TKDE, TNNL, VLDB Journal, and ACM TOIS) with an H-index of 80 and received the prestigious IEEE Computer Society’s AI’s 10 to Watch Award, AIPS Young Tall Poppy Science Award, ARC Future Fellowship 2021 and DECRA Award 2016.
\end{IEEEbiography}

\clearpage
\section{Appendix}
\subsection{Leiden Community Detection Algorithm}
The Leiden algorithm~\cite{Traag2019-51} is an advanced graph clustering technique developed to address limitations of the Louvain method by guaranteeing well-connected communities without compromising computational efficiency. The algorithm operates in three key phases: (1) local moving, where nodes are iteratively reassigned to optimize modularity; (2) refinement, which ensures community connectivity through partitioning; and (3) aggregation, where the network is coarsened for hierarchical clustering. This algorithm is widely used, and its open-source Python implementation is well-documented and accessible. We present the algorithm for Leiden algorithm in Algorithm~\ref{algo:leiden}. 

\begin{algorithm}
	\caption{Leiden Community Detection}
	\label{algo:leiden}
	\SetAlgoLined
	\KwIn{Graph $G(V,E)$, resolution parameter $\gamma$, max iterations $t_{\max}$,
		$\sum_{\text{in}}$: internal edge weights in $\vec{C}$, $\sum_{\text{tot}}$: total edge weights of $\vec{C}$,
		$k_v$: degree of $v$.}
	\KwOut{Community partition $\vec{C}$.}
	
	\SetKwFunction{LEIDEN}{LEIDEN}
	
	Initialize: Partition $\vec{C} = \{\{v\} | v \in V\}$, $t=0$, $Q=-\infty$;\\
	\While{$t < t_{\max}$ \textbf{or} $\Delta Q = \frac{\sum_{\text{in}} + k_{v,\text{in}}}{2m} - \gamma \cdot \frac{(\sum_{\text{tot}} + k_v) \cdot k_v}{(2m)^2} < \epsilon$}{
		\tcp{Phase 1: Local Moving}
		\For{$v \in V$ {\rm in  random order}}{
			$\vec{C}_{\text{old}} = \vec{C}(v)$;\\
			$\vec{C}_{\text{new}} = argmax_{\vec{C}} \Delta Q(v \rightarrow \vec{C})$;\\
			\tcp{$\Delta Q = \frac{\sum_{\text{in}} + k_{v,\text{in}}}{2m} - \gamma \cdot \frac{(\sum_{\text{tot}} + k_v) \cdot k_v}{(2m)^2}$}
			\If{$\Delta Q(v \rightarrow \vec{C}_{\text{new}}) > 0$}{
				$\vec{C}.\text{move}(v, \vec{C}_{\text{new}})$;\\
				Update $\sum_{\text{in}}$, $\sum_{\text{tot}}$ for $\vec{C}_{\text{old}}$ and $\vec{C}_{\text{new}}$;\\
			}
		}
		
		\tcp{Phase 2: Refinement}
		\For{$\vec{C}_k \in \vec{C}$}{
			Split $\vec{C}_k$ into connected components $\{\vec{C}_k^1, \vec{C}_k^2, \ldots\}$;\\
			Merge subsets with $\Delta Q = \frac{\sum_{\text{in}} + k_{v,\text{in}}}{2m} - \gamma \cdot \frac{(\sum_{\text{tot}} + k_v) \cdot k_v}{(2m)^2} > 0$;\\
		}
		
		\tcp{Phase 3: Aggregation}
		$G, \vec{C} \leftarrow \text{aggregate\_network}(G, \text{refined\_partition})$;\\
		$t = t + 1$;\\
		\Return{$\vec{C}$}
	}
\end{algorithm}

\subsection{MapReduce Process}
We provide pseudocode in Algorithm~\ref{algo:map_reduce} to show how GraphRAG utilizes the MapReduce pattern to process large-scale knowledge graphs in parallel and generate more comprehensive responses through distributed computing. The Map phase extracts relevant information from different portions of the knowledge graph, while the reduce phase synthesizes these partial results. 
In practice, we employ the LlamaIndex library (https://docs.llamaindex.ai/en/stable/) to perform retrieval and generation tasks in GraphRAG, including the MapReduce process for generating answers in response to symptom description queries $x$.

\begin{table}
	\centering
	\caption{Inference FLOPs for each model. }
	\label{tb:inference_flops}
	\small
	\begin{tabular}{|p{2.4cm}|p{1.4cm}|p{1.4cm}|p{1.4cm}|p{0.001cm}}
		\toprule
		\centering Model &\centering k=1&\centering k=2& \ \ \ \  k=3  \\ 
		\midrule 
		\centering LLaMA & \centering \textbf{$\sim$1.21$\cdot\textbf{10}^{\textbf{6}}$T}&\textbf{$\sim$1.21$\cdot\textbf{10}^{\textbf{6}}$T}& \textbf{$\sim$1.21$\cdot\textbf{10}^{\textbf{6}}$T}\\ 
		\centering Qwen  & \textbf{$\sim$1.21$\cdot\textbf{10}^{\textbf{6}}$T}&\textbf{$\sim$1.21$\cdot\textbf{10}^{\textbf{6}}$T}	&\textbf{$\sim$1.21$\cdot\textbf{10}^{\textbf{6}}$T}\\ 
		\centering Kimi  &\textbf{$\sim$1.21$\cdot\textbf{10}^{\textbf{6}}$T}&	\textbf{$\sim$1.21$\cdot\textbf{10}^{\textbf{6}}$T}&	\textbf{$\sim$1.21$\cdot\textbf{10}^{\textbf{6}}$T} \\ 
		\centering DeepSeek &\textbf{$\sim$1.21$\cdot\textbf{10}^{\textbf{6}}$T}&\textbf{$\sim$1.21$\cdot\textbf{10}^{\textbf{6}}$T}&\textbf{$\sim$1.21$\cdot\textbf{10}^{\textbf{6}}$T}\\  \midrule
		\centering TCMLLM   &\textbf{$\sim$1.21$\cdot\textbf{10}^{\textbf{6}}$T} &\textbf{$\sim$1.21$\cdot\textbf{10}^{\textbf{6}}$T}&\textbf{$\sim$1.21$\cdot\textbf{10}^{\textbf{6}}$T}\\  \midrule
		\centering RAG   &\centering  $\sim$1.26$\cdot10^{6}$T& \centering $\sim$1.32$\cdot10^{6}$T &$\sim$1.37$\cdot10^{6}$T \\ 
		\centering GraphRAG &$\sim$1.27$\cdot10^{6}$T &$\sim$1.32$\cdot10^{6}$T & $\sim$1.68$\cdot10^{6}$T\\  \midrule
		\centering RAG+Fine-tuning&\centering $\sim$1.26$\cdot10^{6}$T & $\sim$1.32$\cdot10^{6}$T& $\sim$1.37$\cdot10^{6}$T   \\ 
		\centering RAG+SFT  &\centering $\sim$1.26$\cdot10^{6}$T & $\sim$1.32$\cdot10^{6}$T& $\sim$1.37$\cdot10^{6}$T \\ \midrule
		\centering GraphRAG+SFT& \centering $\sim$1.27$\cdot10^{6}$T &$\sim$1.32$\cdot10^{6}$T& $\sim$1.68$\cdot10^{6}$T\\ \midrule
		\centering  \textbf{\RC}& \centering $\sim$1.27$\cdot10^{6}$T & $\sim$1.32$\cdot10^{6}$T& $\sim$1.68$\cdot10^{6}$T \\
		\bottomrule
	\end{tabular}
\end{table}

\subsection{Case Study}
To evaluate the advantages of the TCM formula explanations generated by \sRC compared to other competitive baseline methods, we randomly selected two symptoms from both the collected dataset and the clinical dataset. The models are then tasked with providing responses. The two symptom cases and their corresponding results are presented below:
\begin{itemize}
	\item Symptoms case 1 (Clinical Data): based on the patient's symptoms, recommend TCM formulas and provide relevant information. The patient reports persistent back fatigue, palpitations, and shortness of breath over a 10-month period. These symptoms exhibit an intermittent pattern, occurring both during emotionally stimulating events and at rest. Notably, the patient has no prior history of hypertension, diabetes, or cervical spondylosis. Despite a lifelong engagement in physical activity, these symptoms have only manifested recently.
	
	\item Symptoms case 2 (Collected Data): based on the patient's symptoms, recommend TCM formulas and provide relevant information. The patient's presentation—including chest and diaphragmatic distension, postprandial epigastric oppression, low-grade fever, extremity pain, progressive weight loss, reduced appetite, and generalized weakness—a Classic Ancient Formula is recommended under Traditional Chinese Medicine (TCM) principles, with further diagnostic and therapeutic considerations to be specified.
	
\end{itemize}

\noindent We select DeepSeek, TCMLLM,  RAG, GraphRAG, GraphRAG+SFT and~\RC, and output corresponding explanations.  \textbf{Notably, the explanations are originally in Chinese, but for ease of reading and understanding, we translate it into English. Besides, for clarity, the model’s outputs are color-coded: correct passages appear in bold black, while errors are marked in bold red.
}
Notably, we  include these methods in Table~\ref{tb:case_study1} and Table~\ref{tb:case_study2} to address symptoms case 1, and Table~\ref{tb:case_study3} and Table~\ref{tb:case_study4} to address symptoms case 2,
comparative analysis demonstrates that: 1)~\sRC generates significantly more comprehensive and clinically accurate responses than baseline models, while exhibiting no observable hallucination in its outputs.
2) DeepSeek can output recommended TCM formulas but fails to provide \textit{fine-grained information} (e.g., tongue diagnosis). Besides, DeepSeek's performance in clinical data is worse, possibly because it has never seen comparable cases during its training period.
3) TCMLLM can only provide the formula's sovereign, minister, assistant, courier but lacks other information. 
4) RAG and GraphRAG sometimes fail to retrieve concise information, which causes hallucinations phenomenon (such as generating non-existent prescriptions).
5) Compared to RAG, GraphRAG can retrieve the context with fine granularity, which leads better performance.
6) Since~\sRC  integrates GraphRAG with fine-tuning techniques for LLMs, the answers provided by~\sRC are more refined and accurate.
7) GraphRAG+SFT's nearly matches~\sRC in overall performance, yet its outputs remain slightly less precise and can still contain hallucinations.

Table~\ref{tb:inference_flops} reports the inference FLOPs for each method with top-$k$ values of 1, 2, and 3. The results demonstrate that~\sRC incurs slightly higher computational overhead during inference compared to directly using the LLM API, particularly for $k$ = 1 and 2. However, given its substantial performance improvement, this marginal increase in computation time is acceptable.

\subsection{Hyper-parameter Analysis}

\begin{figure}
	\centering
	\mbox{
		\begin{subfigure}{0.24\textwidth}
			\centering
			\includegraphics[width=\linewidth]{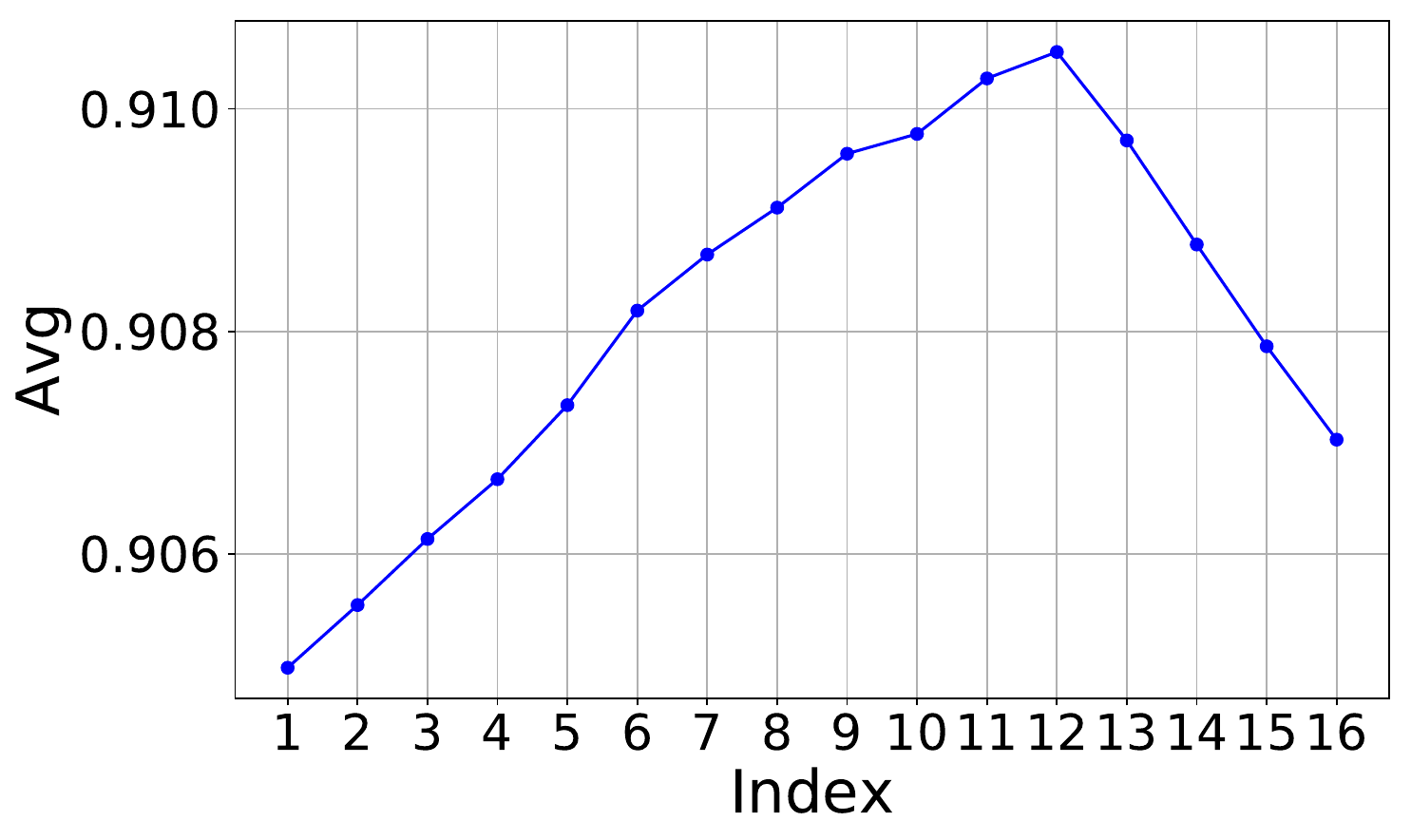}
			\caption{LoRA rank $\tau$}
			\label{subfig:fig1}
		\end{subfigure}%
		\hspace{0.01\textwidth}
		\begin{subfigure}{0.24\textwidth}
			\centering
			\includegraphics[width=\linewidth]{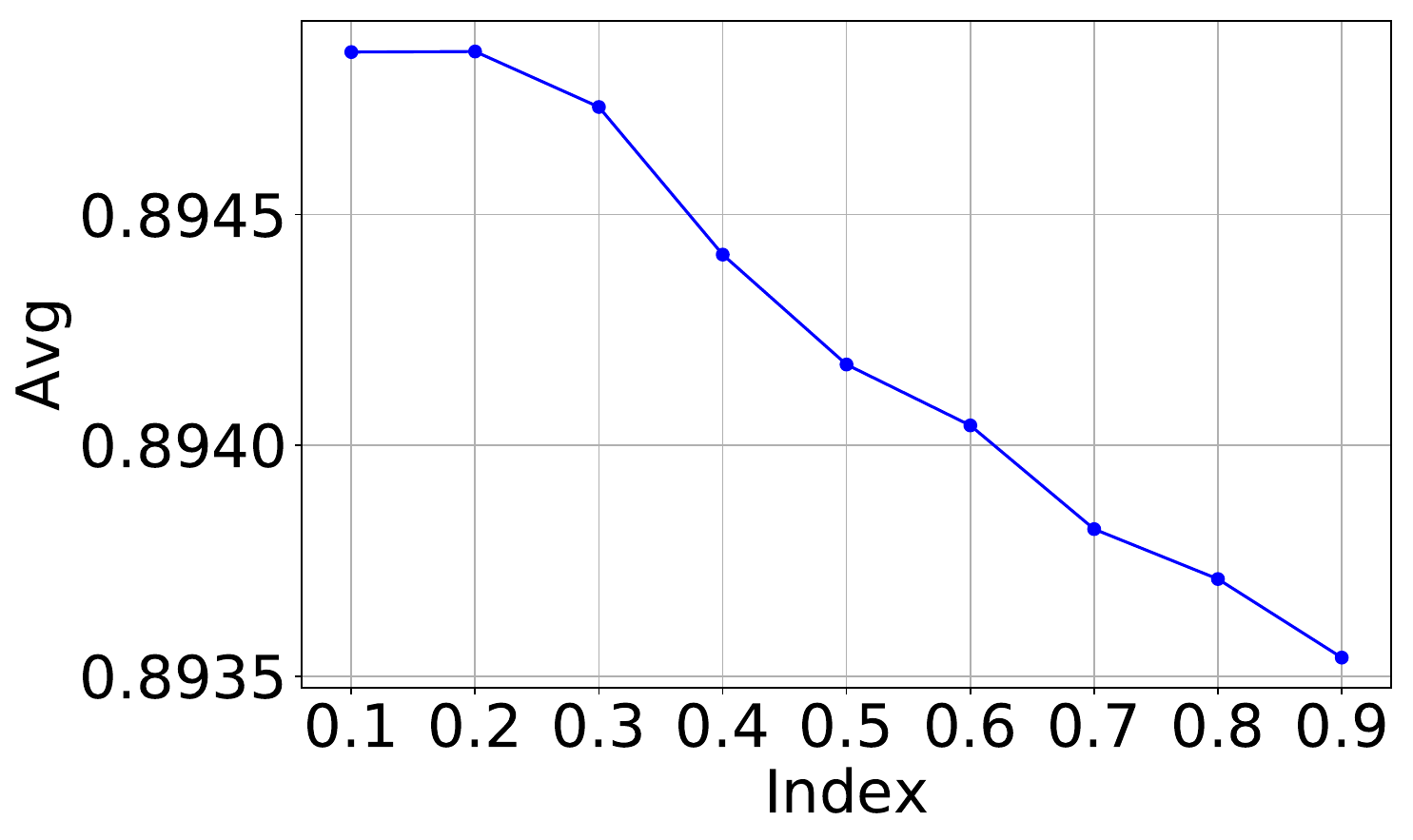}
			\caption{DPO $\beta$}
			\label{subfig:fig2}
		\end{subfigure}
	}
	\caption{Hyper-parameter analysis}
	\label{fig:hyper}
\end{figure}

We conduct a systematic investigation into the impact of two critical hyperparameters—the LoRA rank (denoted as $\tau$) and the DPO key parameter ($\beta$)—on the performance of \RC. The experimental results, presented in Fig.~\ref{fig:hyper}, reveal that: 1)  Optimal performance occurs at $\tau$ = 12, indicating a balance between model capacity and efficiency.2) DPO Parameter ($\beta$): Peak performance is achieved at $\beta$ = 0.2, consistent with prior work (e.g., LLaMA~\cite{Dubey-Arxiv2024-23}).

\begin{algorithm}
	\caption{Map-Reduce Pseudocode for GraphRAG}
	\label{algo:map_reduce}
	\SetAlgoLined
	\KwIn{User query $x$, knowledge graph $\mathcal{G}$, number of shards \textit{num\_shards}}
	\KwOut{Final response to the query}
	
	\SetKwFunction{Map}{map}
	\SetKwFunction{Reduce}{reduce}
	\SetKwFunction{GraphRAG}{graph\_rag\_map\_reduce}

	\textbf{Map\ ({$\textbf{x}$, $\bm{\mathcal{G}}$}):}
		\tcp{Input: User query $x$ and knowledge graph $\mathcal{G}$}
		\tcp{Output: Relevant subgraph $\mathcal{G}_x^*$ and initial response}
		
		\tcp{1. Retrieve relevant nodes from the knowledge graph}
		$\text{relevant\_nodes} \leftarrow \text{retrieve\_nodes}(x, \mathcal{G})$;\\
		
		\tcp{2. Expand to include neighboring nodes forming a subgraph}
		$\text{subgraph} \leftarrow \text{expand\_to\_subgraph}(\text{relevant\_nodes}, \mathcal{G})$; \\
		
		\tcp{3. Perform initial analysis on the subgraph}
		$\text{initial\_analysis} \leftarrow \text{analyze\_subgraph}(\text{subgraph}, x)$;\\
		
		\tcp{4. Generate initial response}
		$\text{initial\_response} \leftarrow \text{generate\_initial\_response}(\text{initial\_analysis})$;\\
		
		\textbf{emit}(\text{subgraph}, \text{initial\_response});\\

	\textbf{Reduce\ ({$\text{subgraphs}$, $\text{initial\_responses}$}):}{
		\tcp{Input: Multiple subgraphs and their corresponding initial responses}
		\tcp{Output: Synthesized final response}
		
		\tcp{1. Merge relevant subgraphs}
		$\text{merged\_graph} \leftarrow \text{merge\_subgraphs}(\text{subgraphs})$;\\
		
		\tcp{2. Perform global graph analysis}
		$\text{global\_analysis} \leftarrow \text{analyze\_merged\_graph}(\text{merged\_graph})$;\\
		
		\tcp{3. Combine with initial responses}
		$\text{combined\_analysis} \leftarrow \text{combine\_analyses}(\text{global\_analysis}, \text{initial\_responses})$;\\
		
		\tcp{4. Generate final response}
		$\text{final\_response} \leftarrow \text{generate\_final\_response}(\text{combined\_analysis})$;\\
		
		\Return{$\text{final\_response}$};\\
		
		\textbf{GraphRAG\ ({$\textbf{x}$, $\bm{\mathcal{G}}$, \textit{num\_shards}}):}{
			\tcp{1. Partition the knowledge graph}
			$\text{graph\_shards} \leftarrow \text{partition\_graph}(\mathcal{G}, \textit{num\_shards})$;\\
			
			\tcp{2. Execute Map phase in parallel}
			$\text{map\_results} \leftarrow \text{parallel\_map}(\text{map}, [x] \times \textit{num\_shards}, \text{graph\_shards})$;\\
			
			\tcp{3. Collect all subgraphs and initial responses}
			$\text{all\_subgraphs} \leftarrow []$\\;
			$\text{all\_initial\_responses} \leftarrow []$;\\
			\For{\text{result} $\in$ \text{map\_results}}{
				$\text{all\_subgraphs}.\text{append}(\text{result.subgraph})$;\\
				$\text{all\_initial\_responses}.\text{append}(\text{result.initial\_response})$;\\
			}
			\tcp{4. Execute Reduce phase}
			$\text{final\_response} \leftarrow \text{reduce}(\text{all\_subgraphs}, \text{all\_initial\_responses})$;\\
			\Return{$\text{final\_response}$};\\
		}
	
	}
\end{algorithm}

\subsection{Base LLM Selection}
\begin{figure}
	\centering
	\includegraphics[width= 0.43 \textwidth]{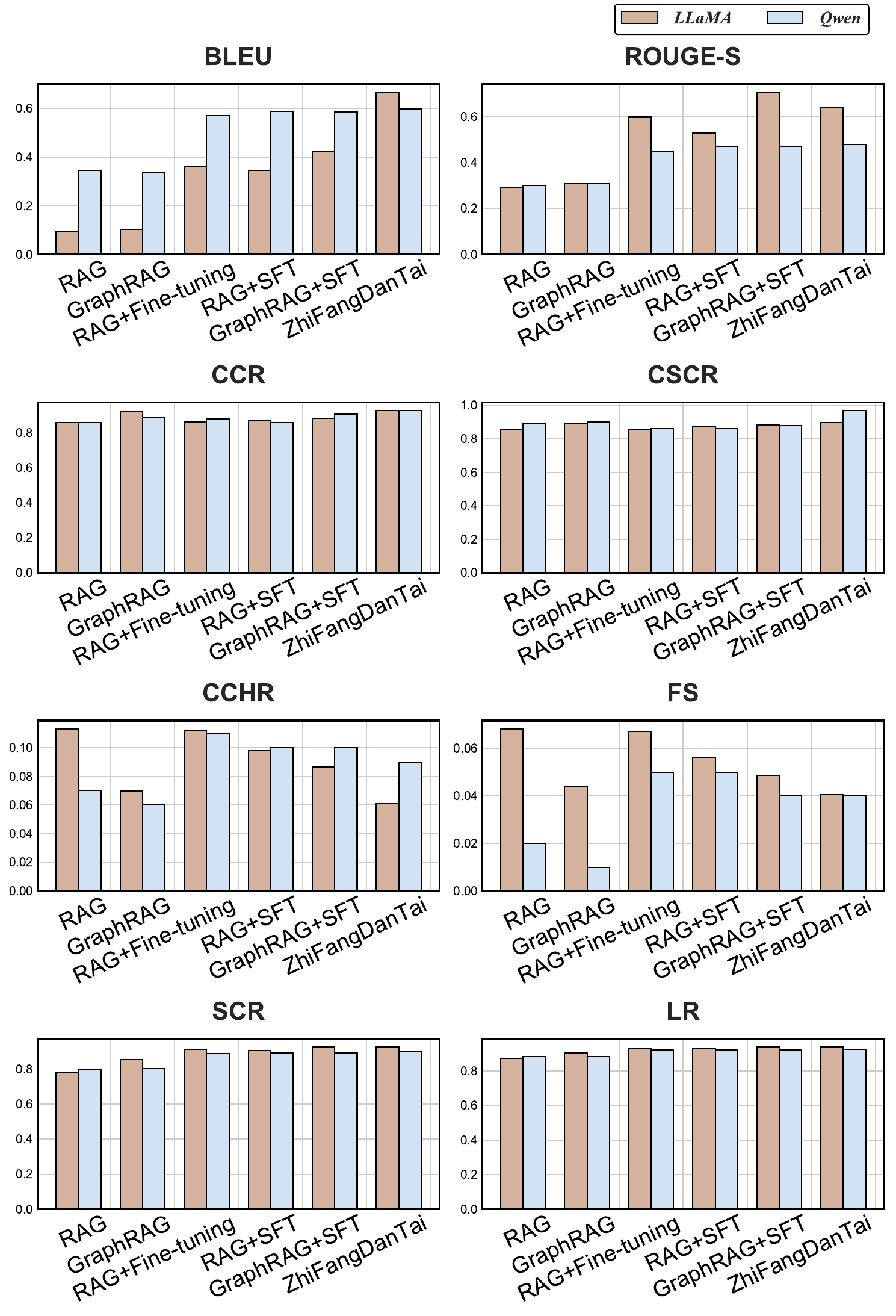}
	\caption{\label{fig:llama_vs_qwen} Performance of using LLaMA and Qwen as base LLM, top-$k$=2 .}
\end{figure}

Furthermore, we compared models based on Qwen and LLaMA, with the results presented in Fig.~\ref{fig:llama_vs_qwen}. The analysis reveals that LLaMA-based models generally outperform Qwen-based models across most metrics, except for BLEU. This discrepancy may arise because LLaMA generates more comprehensive content, albeit with greater divergence from the reference in terms of expression. Although Qwen-based models exhibit marginally better performance in some cases, LLaMA demonstrates superior overall results (see Table~\ref{tb:overall}). Thus, we conclude that selecting a high-performing base model (e.g., LLaMA) is preferable for the TCM formula task.

\subsection{Model Ethical issues}

\subsubsection{Conflict and Absent Knowledge}

To address the issue of conflicting knowledge, we develop specialized instruction datasets containing contradictory information scenarios, including Differences in Medical Theories, Conflicting Information Sources and Practical Problems. These datasets train the model to identify potential conflicts and generate appropriate warning messages, covering the following situations (as shown in Table~\ref{tab:conflicts}).

To address the issue of absent knowledge, as shown in our original paper, i.e., Section-III.D, given a description of symptoms x, we first employ the query expansion technique to obtain the expanded query description $\{x,x'\}$, and then retrieve local answers $\{A_1, \cdots, A_7\}$ based on the fine-grained communities, followed by employing the LLM $\pi_{\theta}$ to obtain the final global retrieval answer $c$. Finally, based on $\{x,x'\}$and $c$, we ask the LLM $\pi_{\theta}$ to generate the TCM formula and corresponding explanations $\mathcal{A}$. 

\subsubsection{Clinical Risks}

To clarify the auxiliary role of~\sRC, we have incorporated the following disclaimer to emphasize that this system serves as a decision-support tool rather than a substitute for clinical judgment:

"Important Note: The prescription recommendations provided are intended for reference purposes only and should not be used without professional supervision. Proper Traditional Chinese Medicine practice requires individualized syndrome differentiation and treatment. For optimal safety and efficacy, please consult a qualified TCM practitioner when using this software."

\subsection{Clinical Data Details}
\begin{figure}
	\centering
	\includegraphics[width=9.8cm,height=6.0cm]{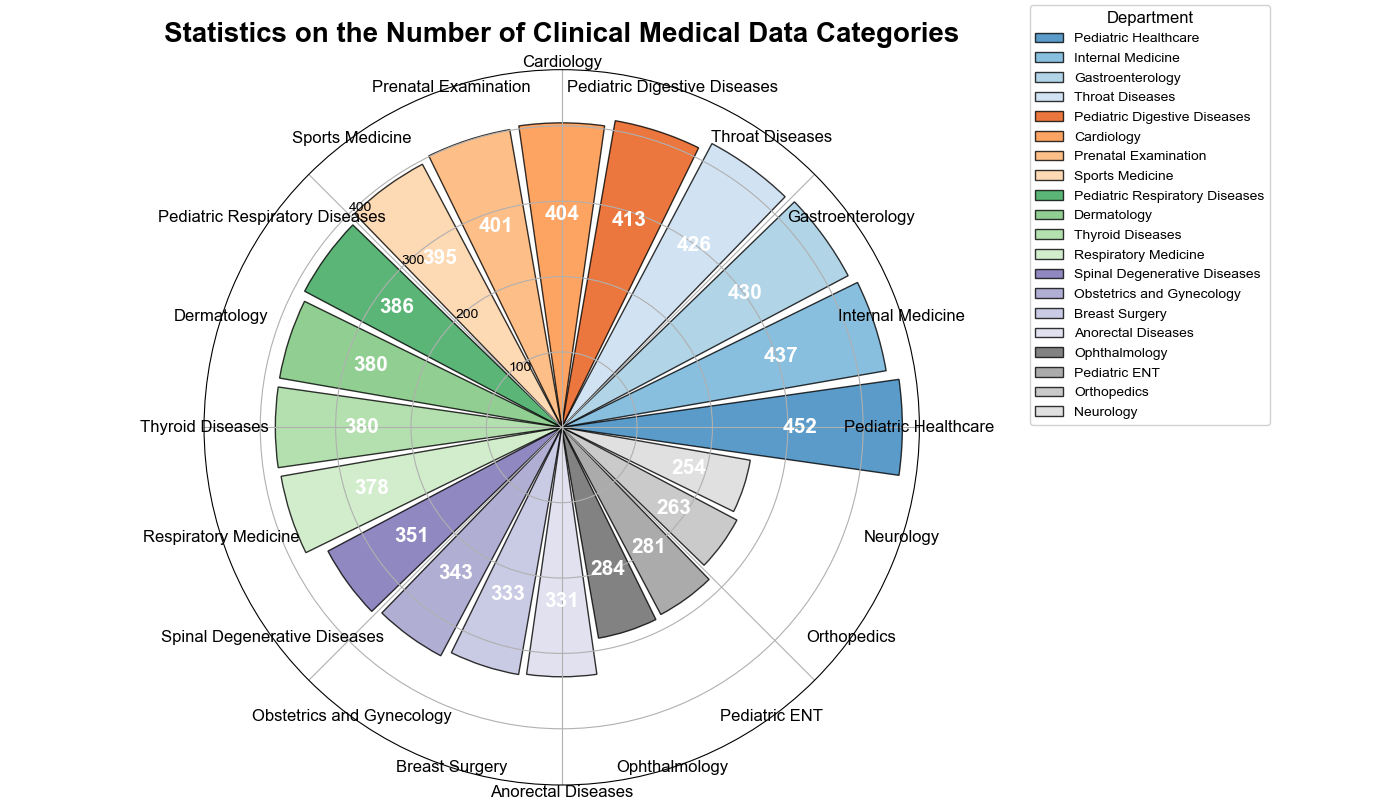}
	\caption{\label{fig:clinical} Classification of department categories.}
\end{figure}

The dataset consists of 30,400 patient records, encompassing a wide range of medical conditions from various specialties. It includes patient demographics, symptom descriptions, diagnostic queries, and specialty classifications, offering valuable insights into prevalent healthcare concerns. Fig.~\ref{fig:clinical} presents the distribution of department categories.

\begin{table*}
	\centering
	\caption{Instruction dataset for contradictory knowledge scenarios}
	\label{tab:conflicts}
	\begin{tabularx}{\linewidth}{|l|>{\raggedright\arraybackslash}X|>{\raggedright\arraybackslash}X|}
		\toprule
		\centering \textbf{Major categories of conflicts} & \centering \textbf{Conflict subclass} & \ \ \ \ \  \ \ \ \ \ \ \ \ \ \ \ \ \ \ \ \ \ \ \ \ \ \textbf{Description} \\
		\midrule
		
		\multirow{16}{*}{Differences in Medical Theories} &\centering  \multirow{8}{*}{TCM vs. Western Medicine} & 
		\begin{itemize}
			\item TCM Approach: For colds ("wind-cold"), TCM uses warming herbs (like Ephedra Decoction) to treat the whole body's balance.
			\item Western Approach: Views colds as viral infections, treating symptoms with medicines like fever reducers.
		\end{itemize} \\
		\cline{2-3}
		
		& \centering \multirow{7}{*}{Differences Within TCM Itself} & 
		\begin{itemize}
			\item Some patients want ancient remedies (like Zhang Zhongjing's formulas), while doctors may prefer newer treatments.
			\item Example: For fever, old-style doctors might use Cinnamon Twig Tea, while modern TCM doctors might choose Forsythia Powder.
		\end{itemize} \\

		\hline
		
		\multirow{16}{*}{Conflicting Information Sources} & \centering \multirow{8}{*}{Old Books vs. New Science} & 
		\begin{itemize}
			\item Old Books: Said some herbs (like cinnabar) were safe.
			\item New Science: Shows these herbs can be toxic with long-term use.
			\item Problem: Safety standards have changed over time.
		\end{itemize} \\
			\cline{2-3}
		
		& \centering \multirow{8}{*}{Regional Differences} & 
		\begin{itemize}
			\item South China: Doctors often use cooling herbs (like poria) for damp-heat conditions.
			\item North China: Doctors prefer warming herbs (like astragalus) for cold conditions.
			\item Issue: Same symptoms get different treatments in different places.
		\end{itemize} \\
		\hline
		
	 \multirow{12}{*}{ Practical Problems} & \centering \multirow{5}{*}{Personal Needs vs. General Advice} & 
		\begin{itemize}
			\item A computer program might suggest ginseng for energy, not realizing it could harm someone with "yin deficiency."
		\end{itemize} \\
			\cline{2-3}
		
		& \centering \multirow{7}{*}{Laws vs. Medical Practice} & 
		\begin{itemize}
			\item China: Allows processed aconite (a poisonous plant used carefully in medicine).
			\item West: Bans aconite because it's dangerous.
			\item Challenge: Computer programs need to know where users live to give proper advice.
		\end{itemize} \\

		\bottomrule
	\end{tabularx}
\end{table*}

\begin{table*}
	\centering
	\caption{Responses to symptoms case 1 (Clinical Data) by DeepSeek, TCMLLM and RAG.}
	\label{tb:case_study1}
	\small
	\begin{tabular}{p{2.6cm}|p{9.2cm}|p{5.2cm}p{0.001cm}}
		\toprule
		\centering Model & \centering Answer & \centering Pros and Cons& \\ 
		\hline
		\centering\multirow{30}{*}{DeepSeek} & The recommended therapeutic approach is as follows:
		
		\textbf{I. TCM Syndrome Analysis:}
		
		Core pathogenesis involves heart-spleen deficiency from chronic mental strain or constitutional weakness, manifesting as palpitations (heart qi/blood insufficiency) and muscular fatigue (spleen deficiency). Concurrent liver qi stagnation may transform into fire disturbing heart spirit, with potential mild blood stasis (though without stabbing pain or purple tongue, stasis signs remain insignificant). 	Key differential diagnosis requires exclusion of organic heart disease (arrhythmia, ischemia) via ECG/echocardiography. Absent structural pathology suggests functional palpitations or depression syndrome in TCM terms.
		
		\textbf{II. Recommended Interventions:}
		
		Patent medicines: \textbf{\textcolor{red}{Guipi Wan (heart-spleen tonification)}}: Astragalus, Codonopsis, Atractylodes, \textbf{\textcolor{red}{Xiaoyao Wan (liver-soothing)}}: Bupleurum, Angelica, White Peony, \textbf{\textcolor{red}{Wenxin Granules (qi-yin supplementation with mild blood activation)}}. \textbf{\textcolor{red}{Customized decoction: Modified Chaihu Shugan San + Guipi Tang base formula}} (Bupleurum 10g, White Peony 12g, Cyperus 10g, Astragalus 15g, etc.), with additions for yin deficiency (Ophiopogon 10g) or pronounced fatigue (Pueraria 15g).
		
		\textbf{III. Adjunctive Therapies:}
		
		\textbf{\textbf{\textcolor{red}{Lifestyle: Stress reduction (meditation), gentle exercise (Tai Chi).
		Acupressure: PC6 (palpitation relief), CV17 (chest regulation).
		Dietary therapy: Lotus seed + lily bulb congee.}}}
		
		\textbf{IV. Precautions:}
		
		Seek immediate care for worsening symptoms (chest pain, syncope).	
		Re-evaluate prescription if no improvement within 2 weeks.		
		Avoid stimulants (coffee, strong tea) & \textbf{\textbf{Pros:}}
		
		(1) Precise syndrome differentiation: The symptoms are accurately categorized as "palpitations," "chest obstruction," "liver qi stagnation," and "heart-spleen deficiency," which aligns closely with TCM theory.

		\textbf{\textcolor{red}{Cons:}} 
		
		(1) DeepSeek fails to give fine-grained information.
		
		(2) Hallucination issues: For instance, the formula incorrectly includes "Wenxin Granules" which appears to be a hallucinated term.
		
	    (3) Lack of punctuation: The response omits standard punctuation, resulting in high information density that may challenge lay readers while remaining suitable for professionals.
		
		(4) Generic formula selection: While demonstrating strong general applicability (e.g., Guipi Wan, Xiaoyao Wan), the recommendations lack distinctive personalization for individual cases.
		
		(5) Undifferentiated herb composition: The formula components are presented without granular classification according to the traditional "monarch-minister-assistant-guide" hierarchy or detailed pharmacological analysis.
		\\
		\hline
		\centering \multirow{14}{*}{TCMLLM} & The herbal formula combines \textbf{\textcolor{red}{Bupleurum root (Chai Hu)}} and \textbf{\textcolor{red}{white peony root (Bai Shao)}} with \textbf{\textcolor{red}{nutgrass rhizome (Xiang Fu)}} to regulate liver qi and relieve stagnation, while \textbf{\textcolor{red}{astragalus root (Huang Qi), codonopsis root (Dang Shen), and atractylodes rhizome (Bai Zhu)}} work synergistically to tonify the heart and spleen. For calming the spirit and alleviating palpitations, the formula incorporates \textbf{\textcolor{red}{zizyphus seed (Suan Zao Ren)}} and \textbf{\textcolor{red}{polygala root (Yuan Zhi)}}, complemented by \textbf{\textcolor{red}{salvia root (Dan Shen)}} and \textbf{\textcolor{red}{licorice root (Gan Cao)}} to activate blood circulation and harmonize the formula. Clinical modifications include adding ophiopogon tuber \textbf{\textcolor{red}{(Mai Dong)}} and \textbf{\textcolor{red}{schisandra berry (Wu Wei Zi)}} to nourish yin when patients present with dry mouth or including \textbf{\textcolor{red}{pueraria root (Ge Gen) and cinnamon twigs (Gui Zhi)}} to relax tendons and promote yang circulation in cases of pronounced back fatigue. & \textbf{\textcolor{red}{Cons:}}
		
		 It fails to include crucial dosage information or explain the pharmacological rationale behind each component's inclusion, potentially creating confusion for practitioners seeking to implement this treatment protocol effectively. This omission of quantitative parameters and explanatory mechanisms significantly reduces the formula's practical utility and reproducibility in clinical settings.
		
		 \\
		\hline
		\centering \multirow{23}{*}{RAG} & In TCM theory, symptoms such as heart failure or other cardiac concerns may be associated with patterns of excessive heart fire, qi deficiency, or kidney yin deficiency. The following herbal recommendations, which should be used under the guidance of a licensed TCM practitioner, are categorized by therapeutic function: \textbf{\textcolor{red}{(1) Qi-tonifying and yin-nourishing herbs}}, including astragalus (enhances qi and spleen function), prepared rehmannia (supports kidney yin and heart-spleen balance), fragrant solomonseal (clears heart heat and promotes fluid production), and alisma (clears heat and drains liver fire); \textbf{\textcolor{red}{(2) Heart-fire-regulating herbs}}, such as poria (reduces heart fire and dampness), salvia root (nourishes blood and resolves stasis), chrysanthemum (clears heart heat and promotes diuresis), and moutan bark (clears toxins and supports urination); \textbf{\textcolor{red}{(3) Qi-blood-harmonizing herbs}}, including angelica root (nourishes and activates blood), atractylodes (strengthens spleen qi), peony root (resolves blood stasis), and licorice (harmonizes formulas and alleviates spasms); and \textbf{\textcolor{red}{(4) Emotion-regulating herbs}}, such as poria (stabilizes mood), gentian root (clears heart fire), and chrysanthemum. A tailored treatment plan should be developed based on individual constitution and symptom presentation, integrating lifestyle modifications like stress reduction, balanced exercise, and dietary adjustments. \textbf{Medical evaluation is advised if symptoms persist or worsen.} 
		
		&		\textbf{Pros:}
				(1) Explores key TCM theories, focusing on how conditions like qi deficiency, yin deficiency, and excessive heart fire are diagnosed.
		
		(2) Lists how each herb works, helping clarify their medicinal effects and how they’re used in practice.
		
		(3) Stresses professional guidance, advising that treatments should be tailored to each patient’s needs by a qualified TCM practitioner.
		
		 \textbf{\textcolor{red}{Cons}:}
		
		(1) Unstructured formula recommendations - Only provides herb lists without systematic formulation principles or hierarchical organization.
		
		(2) Vague diagnostic basis - Lacks clear syndrome differentiation, making clinical application difficult.
		
		(3) Inaccurate herb attributions - Misrepresents traditional uses (e.g., Alisma/"Ze Xie" for liver fire drainage rather than its primary dampness-clearing function). & \\ \hline
	\end{tabular}
\end{table*}

\begin{table*}
	\centering
	\caption{Responses to symptoms case 1 (Collected Data) by GraphRAG, GraphRAG+SFT and~\RC.}
	\label{tb:case_study2}
	\small
	\begin{tabular}{p{2.6cm}|p{9.2cm}|p{5.2cm}p{0.001cm}}
		\toprule
		\centering Model & \centering Answer & \centering Pros and Cons& \\ 
		\hline
		\centering  \multirow{16}{*}{GraphRAG} & TCM Approach for Your Symptoms (back fatigue/palpitations/shortness of breath):
		
		 \textbf{(I)} \textbf{\textcolor{red}{In TCM, these may indicate heart deficiency, lung qi imbalance, or kidney essence weakness;}}
		 
		  \textbf{(II) Recommended formulas (all require professional prescription):} \textbf{1) Si Wu Tang (blood/qiharmonizer):} Angelica, Safflower - eases palpitations. \textbf{2) Ba Zhen Tang (multi-deficiency tonic):} Ginseng, Astragalus - boosts energy. \textbf{3) Astragalus-Codonopsis (qi booster)}: Improves fatigue. \textbf{4) Ginseng-Astragalus (spleen strengthener):} Enhances stamina Honeysuckle (heat-clearer): For stress-triggered palpitations. \textbf{\textcolor{red}{5) Jin Yin Hua Qing Re Tang}}  : If you experience palpitations, especially when you are emotionally agitated, you may need heat-clearing and detoxifying medications. 
		
		\textbf{(III) Important:} Always consult a licensed TCM practitioner for personalized treatment combining herbs, diet and lifestyle adjustments. & \textbf{\textbf{Pros:}}
		
		(1) The recommendations demonstrate comprehensive TCM knowledge by including multiple classic formulas (Si Wu Tang, Ba Zhen Tang, Huang Qi Dang Shen Tang, etc.).
		(2) Maintain therapeutic balance by covering qi-tonifying, blood-regulating, and heat-clearing herbs

		\textbf{\textcolor{red}{Cons:}} 
		
		(1) Lacks precise syndrome differentiation  (e.g., no clear distinction between qi deficiency, heart yin deficiency, or liver stagnation) 
		(2) Hallucination Issues: Some suggested formulas like "Jin Yin Hua Qing Re Tang" are non-standard combinations.

		\\
		\hline
		\centering \multirow{23}{*}{GraphRAG+SFT} &
		\textbf{(1) Herbal Composition:} Monarch herbs: Ginseng (10g, replenishes qi and calms mind), Salvia (15g, activates blood and clears heart heat);
		Minister herbs: Schisandra (6g, stabilizes heart qi), Ophiopogon (12g, nourishes yin), Asparagus root (10g, reduces fire);
		Assistant herbs: Angelica (10g, nourishes blood), Polygala (6g, tranquilizes mind), Zizyphus seed (15g, calms spirit), Biota seed (10g, moistens intestines);
		Envoy herbs: Platycodon (6g, \textbf{\textcolor{red}{directs herbs upward})}, Honey-fried licorice (6g, harmonizes formula);	
	
		\textbf{(2) Indications:} For heart yin deficiency with deficient fire symptoms - intermittent palpitations, chest tightness, back fatigue triggered by emotional stress, accompanied by spontaneous sweating, dry mouth, and insomnia.
		
		\textbf{(3) Diagnostic Signs:}	Pulse: Thin and rapid or irregular;Tongue: Red with little/no coating
		
		\textbf{(4) Contraindications:}
		Avoid in patients with spleen-stomach cold/diarrhea;
		Incompatible with Veratrum nigrum (antagonistic pairing);
		Dietary restrictions: No spicy foods;
		
		\textbf{(5) Preparation:}
		Powder ginseng separately;
		Grind other herbs into fine powder (100-mesh sieve);
		Form into 9g honey pills with cinnabar coating (modern substitutes available);
		
		\textbf{Dosage:} 1 pill twice daily with warm water
		
		\textbf{Clinical Note:} For athletes with palpitations, rule out organic heart disease with ECG & \textbf{\textbf{Pros:}}
		
		(1) The diagnosis is clearly defined as "heart yin deficiency with deficient fire disturbing the spirit," demonstrating precise syndrome differentiation;
		
		 (2) The herbal composition follows a well-structured monarch-minister-assistant-envoy framework, with comprehensive annotations explaining each herb's therapeutic role, ensuring logical formula design and clinical applicability.

		\textbf{\textcolor{red}{Cons:}}
    	(1) Hallucination Issues: Some suggested point (e.g.,``directs herbs upward" ) is incorrect.
		\\
		\hline
		\centering \multirow{27}{*}{~\sRC} & 
		\textbf{(1) Herbal Composition:}
		\textbf{Monarch herbs:} Ginseng (10g, replenishes vital energy and calms palpitations), Salvia root (15g, activates blood circulation and clears heart heat);
		\textbf{Minister herbs:} Schisandra berry (6g, stabilizes heart energy), Ophiopogon tuber (12g, nourishes yin and clears heart fire), Asparagus root (12g, reduces fire and generates fluids);
		\textbf{Assistant herbs:} Angelica root (10g, nourishes blood and softens liver), Polygala root (6g, calms mind and connects heart-kidney), Zizyphus seed (15g, soothes liver and stops sweating), Biota seed (10g, tranquilizes mind and moistens intestines);
		\textbf{Envoy herbs:} Honey-fried licorice (6g, harmonizes formula), Poria (12g, strengthens spleen and calms mind)
		
		\textbf{(2) Indications:} This pattern manifests as heart yin deficiency with internal heat, characterized by palpitations, dyspnea, chest constriction, sleep disturbances, and heightened startle response. 
		
		\textbf{(3) Diagnostic Signs:}
		Pulse: Thin-rapid or wiry-thin;
		Tongue: Red with little/no coating, possibly cracked;
		
		\textbf{(4) Contraindications:}
		Not recommended for those with cold spleen/stomach or loose stools; use caution in pregnancy;
		Avoid with Veratrum and Trogopterus dung (incompatible combinations);
		Precautions: Avoid spicy/greasy foods; minimize stress and overexertion
		
		\textbf{(5) Preparation:}
		Powder ginseng separately;
		Decoct other herbs (except Poria) twice, concentrate to thick paste;
		Mix with ginseng powder to form pills (6-9g/dose, twice daily);
		Alternative: Prepare as decoction (1 dose/day, divided);
		
		\textbf{Important:} Important Note: The prescription recommendations provided are intended for reference purposes only and should not be used without professional supervision.
		
		&		\textbf{\textbf{Pros:}} 
		
		(1) The formulation aligns with and improves upon the GraphRAG+SFT approach through refined therapeutic descriptions; 
		
		(2) It demonstrates exceptional structural standardization, complete with properly categorized monarch-minister-assistant-envoy herbs, detailed tongue/pulse diagnostics, contraindications, preparation methods, and administration guidelines; 
		
		(3) The design exhibits strong systematic capability through standardized formula presentation and enhanced clinical usability. 
		
		(4) Hallucination Analysis: The formula maintains high reliability with near-zero hallucinations, featuring correct herb combinations and fully compliant composition standards.
		& \\ \hline
	\end{tabular}
\end{table*}

\begin{table*}
	\centering
	\caption{Responses to symptoms case 2 (Collected Data) by DeepSeek, TCMLLM and RAG.}
	\label{tb:case_study3}
	\small
	\begin{tabular}{p{2.6cm}|p{9.2cm}|p{5.2cm}p{0.001cm}}
		\toprule
		\centering Model & \centering Answer & \centering Pros and Cons& \\ 
		\hline
		\centering\multirow{31}{*}{DeepSeek} & \textbf{Recommendation:} Modified Chaihu Guizhi Tang (Bupleurum-Cinnamon Combination):
		
		\textbf{\textcolor{red}{Composition}} \textbf{(dose adjustment required per constitution):}
		Bupleurum (liver regulation);
		Scutellaria (heat clearance);
		Cinnamon twig (channel warming);
		White peony (blood nourishment);
		Codonopsis (qi tonification);
		Atractylodes  (dampness resolution);
		Poria (spleen fortification);
		Citrus peel  (qi regulation);
		Pinellia (phlegm transformation);
		Ophiopogon (yin nourishment);
		Honey-fried licorice (formula harmonization);
		
		\textbf{Dosage adjustment:}
		Marked false fever: Add Lycium bark , Stellaria root;
		Severe distension: Add Aucklandia, Amomum (added last);
		Profound emaciation: Add Dioscorea, \textbf{\textcolor{red}{Polygonatum;}}
		
		\textbf{Therapeutic Rationale:}
		Liver-spleen regulation (Bupleurum+Peony for distension; Codonopsis+Atractylodes+Poria for digestive complaints);
		Heat clearance (Scutellaria+Ophiopogon for fever/extremity discomfort)
		Qi-blood restoration (Cinnamon for yang assistance; Citrus+Pinellia for qi-phlegm dynamics)
		
		\textbf{Patent Medicine Alternatives:}
		Xiaoyao Wan (liver-spleen regulation) + Shengmai Yin (qi-yin supplementation) for mild cases;
		Yueju Baohe Wan (qi stagnation relief) + Renshen Jianpi Wan (spleen strengthening) for predominant stagnation;
		
		\textbf{Adjunctive Management:}
		Diet: Avoid raw/cold foods; consume small frequent meals (Dioscorea congee, Citrus tea);
		Lifestyle: Moderate exercise (e.g., Baduanjin); stress reduction;
		Acupressure: PC6 (chest discomfort), ST36 (spleen support), LV3 (liver regulation);
		
		\textbf{Clinical Precautions:}
		Rule out organic pathologies (chronic gastritis, thyroid dysfunction);
		Re-evaluate if no improvement within 1 week;
		Caution with drying herbs in yin-deficient heat presentations (red tongue, tidal fever);
		
		\textbf{Professional TCM consultation is essential for personalized dosage and syndrome differentiation.}
		
	& \textbf{\textbf{Pros:}}
		
		(1) comprehensive syndrome differentiation by accurately identifying "consumptive fatigue with qi stagnation" and correctly deriving the pathogenesis as "liver-spleen disharmony with qi-yin deficiency." 
		 
		 (2) providing clinically valuable adjunct recommendations including patent medicine alternatives, dietary guidance, and acupressure points.

		\textbf{\textcolor{red}{Cons:}} 
		
		(1) potential information overload due to text density, which may challenge users with limited information integration capacity.
		
		(2) absence of explicit monarch-minister-assistant-envoy herb classification. 
				 
		(3) Lack of drug dosage.
		
		(4) Few hallucination instances, e.g., ``Polygonatum".
	
		\\
		\hline
		\centering \multirow{12}{*}{TCMLLM} &  \textbf{\textcolor{red}{	Bupleurum 12g (liver regulation);
		Scutellaria 9g (heat clearance);
		Cinnamon twig 9g (channel warming);
		White peony 12g (blood nourishment);
		Codonopsis 15g (qi tonification);
		Atractylodes 12g (dampness resolution);
		Poria 15g (spleen fortification);
		Citrus peel 9g (qi regulation);
		Pinellia 9g (phlegm transformation);
		Ophiopogon 12g (yin nourishment);
		Honey-fried licorice 6g (formula harmonization);}}
		& \textbf{\textcolor{red}{Cons:}}
		
		 It fails to include crucial dosage information or explain the pharmacological rationale behind each component's inclusion, potentially creating confusion for practitioners seeking to implement this treatment protocol effectively. This omission of quantitative parameters and explanatory mechanisms significantly reduces the formula's practical utility and reproducibility in clinical settings.

		\\
		\hline
		\centering \multirow{13}{*}{RAG} &Medication Protocol for Zhi Shi Wan (Aurantii Immaturus Pill): This powdered formula (\textbf{\textcolor{red}{composed of Immature Bitter Orange, Ripe Bitter Orange, and Licorice)}} is administered orally with water at a dosage of 20 pills per dose, progressively increasing to 30 pills, taken twice daily on an empty stomach. It is clinically indicated for managing symptoms including thoracic distension, appetite loss, false fever, extremity pain, progressive emaciation, and fatigue with reduced food intake. The gradual dosage escalation and fasting administration optimize bioavailability for this qi-regulating formula.
		&		\textbf{\textcolor{red}{Cons:}}
		(1) Significant information gaps exist regarding pharmacological composition details, therapeutic indication analysis, and mechanistic rationale, limiting clinical utility; 
		(2) The formula demonstrates content redundancy with duplicated listing of "Citrus aurantium" (bitter orange), reflecting either documentation errors or formulation inaccuracies that require verification against standard pharmacopeia references.
		(3) Few hallucination instances, e.g., ``Bitter Orange, Ripe Bitter Orange".
		
		 & \\ \hline
	\end{tabular}
\end{table*}

\begin{table*}
	\centering
	\caption{Responses to symptoms case 2 (Collected Data) by GraphRAG, GraphRAG+SFT and~\RC.}
	\label{tb:case_study4}
	\small
	\begin{tabular}{p{2.6cm}|p{9.2cm}|p{5.2cm}p{0.001cm}}
		\toprule
		\centering Model & \centering Answer & \centering Pros and Cons& \\ 
		\hline
		\centering\multirow{22}{*}{GraphRAG} & \textbf{two herbal formulations are recommended:}
		
		\textbf{(1) Aurantii Immaturus Pill (Zhi Shi Wan):}
		\textbf{\textcolor{red}{Composition: Immature bitter orange, rhinoceros horn (note: ethically restricted; alternatives recommended), hogfennel, costus root, ophiopogon tuber, red poria, sophora root, and peony root.}}
		\textbf{Administration:} Initially 20 pills, gradually increasing to 30 pills, taken twice daily on an empty stomach with warm water.
		Preparation: Herbs are ground into powder and formed into honey-fused pills the size of phoenix tree seeds.
		
		\textbf{(2) Bupleurum Liver-Soothing Powder (Chai Hu Shu Gan San):}
		Composition: Bupleurum root, white peony root, bitter orange peel, licorice root, cyperus rhizome, and aged tangerine peel.
		Administration: Three times daily at flexible intervals, with dosage adjusted per clinical evaluation.
		Preparation: Herbs are finely powdered and uniformly blended.
		
		Clinical Considerations:
		\textbf{\textcolor{red}{While both formulations target the aforementioned symptoms}}, their distinct compositions (e.g., inclusion of heat-clearing vs. qi-regulating herbs) and \textbf{preparation methods necessitate individualized selection based on patient-specific pattern differentiation and professional medical guidance.}

		& \textbf{\textbf{Pros:}}
		
		(1) The comparative presentation of Aurantii Immaturus Pill and Bupleurum Liver-Soothing Powder provides reasonable therapeutic alternatives for symptom management

		\textbf{\textcolor{red}{Cons:}} 
		
		(1) Inaccurate composition—the listed Aurantii Immaturus Pill erroneously includes non-standard ingredients (e.g., rhinoceros horn, ophiopogon, and sophora root), deviating from classical formulations; 
		
		(2) Overgeneralized indications—the rationale for addressing thoracic distension and anorexia lacks
	
	    (3) Hallucination Analysis:
	    
	    Formula contamination: The Aurantii Immaturus Pill incorporates extraneous herbs (e.g., heat-clearing agents) inconsistent with its traditional structure.
	    	
		\\
		\hline
		\centering \multirow{20}{*}{GraphRAG+SFT} &  \textbf{(1) Therapeutic Recommendation:} Aurantii Immaturus Pill (Zhi Shi Wan)
		This formula is primarily indicated for patients presenting with chest and diaphragmatic distension, postprandial epigastric discomfort, false fever, extremity pain, progressive emaciation, reduced appetite, and generalized weakness. 
		
		\textbf{(2) The formulation consists of:} (1) \textbf{Monarch herb} - Immature bitter orange (\textbf{\textcolor{red}{6 parts}}, for qi regulation and distension relief); (2) \textbf{Minister herbs} - Rhinoceros horn (\textbf{\textcolor{red}{4 parts}}, heat-clearing), Hogfennel (\textbf{\textcolor{red}{4 parts}}, lung-descending), Costus root (\textbf{\textcolor{red}{8 parts}}, qi-moving and pain-relieving), and \textbf{Ophiopogon tuber} (\textbf{\textcolor{red}{8 parts}}, yin-nourishing); (3) \textbf{Assistant herbs} - Red poria (\textbf{\textcolor{red}{8 parts}}, dampness-eliminating), Sophora root (\textbf{\textcolor{red}{6 parts}}, \textbf{\textcolor{red}{heat-drying}}), and \textbf{Peony root} (\textbf{\textcolor{red}{6 parts}}, blood-nourishing). 
		 \textbf{\textcolor{red}{69 grams each time, 23 times a day. The specific dosage can be adjusted according to the patient's spleen and stomach condition.}}
		 
		\textbf{(3) Diagnostic indicators include} wiry or thin pulse, possibly slippery, with red tongue and yellow coating suggesting internal heat/dampness.
		
		 \textbf{(4) Contraindications include} spleen-stomach cold deficiency and pregnancy, with cautions against concurrent use with warming tonics like ginseng.  Though the inclusion of non-traditional components (e.g., rhinoceros horn) requires modern ethical alternatives.
		& 	 \textbf{\textbf{Pros:}}
		
		(1) clear hierarchical organization with well-defined monarch-minister-assistant-envoy roles and logical structure.
		
		 (2) strong diagnostic elements, incorporating tongue/pulse indicators and patient-specific indications that align with TCM pattern differentiation principles. 
		
		\textbf{\textcolor{red}{Cons:}}
		
	(1) Non-classical composition, as the formula deviates from traditional formulations.
	
	(2) Ambiguous dosing, lacking precise quantitative standards or clinical references.
		
		(3) Few hallucination instances: e.g., the frequency of use and dosage are incorrect
		
		\\
		\hline
		\centering \multirow{26}{*}{~\sRC} &\textbf{(1) This herbal preparation is indicated for patients presenting with} chest and abdominal distension, low-grade fever, extremity discomfort, poor appetite, fatigue, and progressive weight loss—symptoms typically associated with qi stagnation, damp-heat accumulation, or spleen-stomach deficiency. 
		
		\textbf{(2) The formula contains:} 1) Primary agent: immature bitter orange (6 parts, for relieving qi stagnation and distension); 2) Supporting agents: rhinoceros horn (4 parts, heat-clearing), hogfennel (4 parts, lung-descending), and costus root (8 parts, qi-regulating); 3) Adjuvant herbs: ophiopogon tuber (8 parts, yin-nourishing), red poria (8 parts, dampness-resolving), sophora root (6 parts, heat-clearing), and peony root (6 parts, blood-nourishing).
		
		\textbf{(3) Diagnostic indicators include} a wiry or slippery pulse and a red tongue with yellow/white coating. 
		
		\textbf{(4) Contraindications include} pregnancy and cold-type spleen-stomach deficiency. 
		
		\textbf{(5) The preparation involves} grinding herbs into powder and forming pills (6-9g per dose, 2-3 times daily), with dietary precautions against greasy/spicy foods. 
		
		\textbf{(6) Note: Rhinoceros horn requires substitution with ethical alternatives in modern practice.}
		
		\textbf{Important Note}: The prescription recommendations provided are intended for reference purposes only and should not be used without professional supervision. Proper Traditional Chinese Medicine practice requires individualized syndrome differentiation and treatment. For optimal safety and efficacy, please consult a qualified TCM practitioner when using this software."
		&		\textbf{{Pros:}}
	
		(1) comprehensive documentation by detailing composition, indications, diagnostic markers, contraindications, preparation methods, and administration guidelines.
	
		(2) effective organization through logical structuring that facilitates systematic review.
	
		(3) Hallucination Analysis: The formula maintains high reliability with near-zero hallucinations, featuring correct herb combinations and fully compliant composition standards.

		& \\ \hline
	\end{tabular}
\end{table*}

\end{document}